\documentclass[journal,onecolumn,draftcls,12pt]{IEEEtran}
\usepackage[letterpaper, left=0.7in, right=0.7in, bottom=0.99in, top=0.7in]{geometry}
\IEEEoverridecommandlockouts

\usepackage{cite}
\usepackage{amsmath,amssymb,amsfonts, mathtools, amsthm}
\usepackage{algorithmic}
\usepackage{graphicx}
\usepackage{textcomp}
\usepackage{xcolor}
\usepackage{booktabs}
\usepackage{glossaries}
\usepackage{algorithm}
\usepackage{algorithmic}
\usepackage{bbm}
\usepackage{bm}
\usepackage{multirow}
\usepackage[normalem]{ulem}
\usepackage{tabularx}
\usepackage{subcaption}
\usepackage{caption}
\usepackage{dsfont}
\newacronym{aoi}{AoI}{Age of Information}
\newacronym{aodv}{AODV}{Ad hoc On-demand Distance Vector}
\newacronym{dsdv}{DSDV}{Destination Sequenced Distance Vector}
\newacronym{cnn}{CNN}{convolutional neural networks}
\newacronym{dl}{DL}{Deep Learning}
\newacronym{dod}{DoD}{depth of discharge}
\newacronym{dqn}{DQN}{deep Q-Learning}
\newacronym{geo}{GEO}{geostationary orbit}
\newacronym{gsl}{GSL}{ground-to-satellite link}
\newacronym{hol}{HOL}{Head of Line}
\newacronym{isl}{ISL}{inter-satellite link}
\newacronym{leo}{LEO}{low Earth orbit} 
\newacronym{ml}{ML}{Machine Learning}
\newacronym{mdp}{MDP}{Markov decision process}
\newacronym{ngeo}{NGSO}{Non-geostationary orbit}
\newacronym{ngso}{NGSO}{non-geostationary orbit}
\newacronym{olsr}{OLSR}{optimized link state routing protocol}
\newacronym{ospf}{OSPF}{Open Shortest Path First}
\newacronym{pan}{PAN}{Path-Aware Networking}
\newacronym{qos}{QoS}{Quality of Service}
\newacronym{rl}{RL}{Reinforcement Learning}
\newacronym{drl}{DRL}{Deep Reinforcement Learning}
\newacronym{ddqn}{DDQN}{Double Deep Q-Learning}
\newacronym{dnn}{DNN}{Deep Neural Network}
\newacronym{dql}{DQL}{Double Q-Learning}
\newacronym{ql}{QL}{Q-Learning}
\newacronym{qr}{QR}{Q-Routing}
\newacronym{td}{TD}{Temporal Difference}
\newacronym{e2e}{E2E}{end-to-end}
\newacronym{bgp}{BGP}{Border Gateway Protocol}
\newacronym{ibgp}{iBGP}{interior Border Gateway Protocol}
\newacronym{ebgp}{eBGP}{exterior Border Gateway Protocol}
\newacronym{as}{AS}{Autonomous System}
\newacronym{relu}{ReLu}{Rectified Linear Unit}
\newacronym{cdf}{CDF}{Cumulative Distribution Function}
\newacronym{gps}{GPS}{Global Positioning System}
\newacronym{pomdp}{POMDP}{Partially Observable Markov Decision Problem}
\newacronym{snr}{SNR}{signal-to-noise ratio}
\newacronym{sgd}{SGD}{Stochastic Gradient Descent}
\newacronym{awgn}{AWGN}{additive-white Gaussian noise}
\newacronym{bm}{BM}{benchmark}
\newacronym{fifo}{FIFO}{first-in first-out}
\newacronym{ip}{IP}{Internet Protocol}
\newacronym{gsd}{GSD}{Ground Sample Distance}
\newacronym{eo}{EO}{Earth Observation}
\newacronym{iot}{IoT}{Internet of Things}
\newacronym{semcom}{SemCom}{Semantic Communications}
\newacronym{sdg}{SDG}{Sustainable Development Goal}
\newacronym{ai}{AI}{Artificial Intelligence}
\newacronym{mse}{MSE}{Mean Square Error}
\newacronym{fid}{FID}{Fréchet Inception Distance}
\newacronym{jscc}{JSCC}{Joint Source and Channel Coding}
\newacronym{kg}{KG}{Knowledge Graph}
\newacronym{gtfp}{GTFP}{ground track frame period}
\newacronym{gs}{GS}{ground station}
\newacronym{fov}{FOV}{field of view}
\newacronym{fso}{FSO}{free-space optical}
\newacronym{ntn}{NTN}{Non-Terrestrial Networks}
\newacronym{lsatc}{LSatC}{Low Earth Orbit Satellite Constellation}
\newacronym{fdrl}{FD-RL}{Federated Deep Reinforcement Learning}
\newacronym{fl}{FL}{Federated Learning}
\newacronym{fedavg}{FedAvg}{Federated Averaging}
\newacronym{sfl}{SFL}{Satellite Federated Learning}
\newacronym{sota}{SoTA}{State of The Art}
\newacronym{ps}{PS}{Parameter Server}
\newacronym{ma-drl}{MA-DRL}{Multi-Agent Deep Reinforcement Learning}
\newacronym{vn}{VN}{Virtual Node}
\newacronym{vt}{VT}{Virtual Topology}
\newacronym{lu}{LU}{Local Update}
\newacronym{los}{LOS}{Line of Sight}
\newacronym{cka}{CKA}{Centered Kernel Alignment}

\newcommand{\ilm}[1]{#1}

\usepackage{balance}

\title{Continual Deep Reinforcement Learning for Decentralized Satellite Routing\vspace{-0.0em}}

\author{\IEEEauthorblockN{Federico Lozano-Cuadra, Beatriz Soret~\IEEEmembership{Senior Member,~IEEE}, Israel Leyva-Mayorga~\IEEEmembership{Member,~IEEE}, Petar Popovski~\IEEEmembership{IEEE Fellow}}
\vspace{-0.0em}
\thanks{F. Lozano-Cuadra (flozano@ic.uma.es) and B. Soret are with the Telecommunications Research Institute, University of Malaga, 29071, Malaga, Spain. I. Leyva-Mayorga and P. Popovski are with the Connectivity Section, Aalborg University, 9220 Aalborg, Denmark. The work of F. Lozano-Cuadra and B. Soret is partially funded by the European Space Agency (ESA) framework SatNEx V (prime contract no. 4000130962/20/NL/NL/FE), and by the Spanish Ministerio de Ciencia, Innovación y Universidades (PID2022-136269OB-I00). The view expressed herein can in no way be taken to reflect the official opinion of ESA.
}
}

\date{}

\def\subparagraph{} 
\usepackage{titlesec}
\titlespacing*{\section}{0pt}{*1}{*1}
\titlespacing{\subsection}{0pt}{*1}{*1}

\renewcommand{\thesubsubsection}{\arabic{subsubsection}}

\titleformat{\subsubsection}[runin]{\itshape}{\thesubsubsection)}{1em}{}
\titlespacing*{\subsubsection}{\parindent}{0pt}{*1}

\begin{document}

\bstctlcite{IEEEexample:BSTcontrol}

\maketitle
\begin{abstract}
This paper introduces a full solution for decentralized routing in \glspl{lsatc} based on continual \gls{drl}. This requires addressing multiple challenges, including the 
partial knowledge at the satellites and their continuous movement, and the time-varying sources of uncertainty in the system, such as traffic, communication links, or communication buffers. We follow a multi-agent approach, where each satellite acts as an independent decision-making agent, while 
acquiring a limited knowledge of the environment based on the feedback received from the nearby agents. The solution is divided into two phases. First, an offline learning phase relies on decentralized decisions and a global \gls{dnn} {trained with global} experiences to learn the optimal paths at each possible position and congestion level. Then, the online phase with local, on-board, and pre-trained \glspl{dnn} requires continual learning to evolve with the environment, which can be done in two different ways: (1) Model anticipation, where the predictable conditions of the constellation, resulting from its orbital dynamics, are exploited by each satellite sharing local model with the next satellite; and (2) \gls{fl}, where each agent's model is merged first at the cluster level and then aggregated in a global \gls{ps} at ground or at a \gls{geo} satellite. The results show that, without high congestion, the proposed \gls{ma-drl} framework achieves the same E2E performance as a shortest-path solution, but the latter assumes intensive communication overhead for real-time network-wise knowledge of the system at a centralized node, whereas ours only requires limited feedback exchange among first neighbour satellites. Importantly, our solution adapts well to congestion conditions and exploits less loaded paths. 
Moreover, the divergence of models over time is easily tackled by the synergy between anticipation, applied in short-term alignment, and \gls{fl}, utilized for long-term alignment.

\end{abstract}
\vspace{-.1cm}

\glsresetall

\section{Introduction}

Through the incremental adoption of the \gls{isl}, \glspl{lsatc} are turning into packet-based \gls{ntn} capable of providing ubiquituous sensing, navigation, positioning, and communication services towards 6G. From a network perspective, \gls{e2e} packet routing in the \gls{lsatc} refers to finding appropriate paths to interconnect  source-destination pairs, where each satellite is a router and the \glspl{gs} are the end-points. This is a complex and dynamic problem with unique characteristics as follows~\cite{soret2023q}\cite{Rabjerg2021}: (1) The topology is dynamic yet predictable, with frequent link disruptions as the satellites move. (2) The propagation delays are high due to the large distances. (3) The injected terrestrial traffic is dynamic, imbalanced and not readily predictable. A space routing solution should strive for being resilient against the various \emph{sources of uncertainty} (links, traffic, buffers) and operate under \emph{partial knowledge} of each individual satellite. In addition, routing should account for the constrained computing and communication resources at the spacecrafts, and the limited connectivity with the ground infrastructure, i.e., the \glspl{gs}. This setup motivates investigation of a learning-based distributed solution to attain a complex balance among performance, system knowledge, uncertainty, and hardware constraints.


Solving the routing problem typically entails modeling the network as a graph, where the nodes are the network elements and the edges are the communication links. Routing in terrestrial networks with stable links can be handled with Dijkstra's algorithm and static routing tables~\cite{dijkstra1959}. However, a centralized optimization of the routes requires global knowledge of the network topology and the nodes and edges weights, e.g., the queues and rates of each link, which has proven to be a complicated and dynamic problem~\cite{Tang2018}. \Glspl{lsatc} add even more dynamics to the routing problem, as satellites can maintain two intra-plane and two inter-plane \gls{isl} with the first neighbours, which might change over time, plus the \gls{gsl} when a \gls{gs} is visible during the pass. Thus, the \gls{lsatc} graph is intrinsically time-variant, and the set of nodes and edges involves satellites and \glspl{gs}, and \glspl{isl} and \glspl{gsl}, respectively. At the same time, space systems typically face stringent computation, communication, and power constraints as compared to terrestrial. While routing in satellite constellations can be solved centrally by assuming perfect knowledge of the queues~\cite{Rabjerg2021}, in practice, neither the ground nor the space segment has real-time knowledge of the dynamic queueing times \emph{and} traffic demands of the whole network. In addition, these can change over time in non-stationary ways, such that it is impractical to work with routing algorithms that rely on global knowledge or strong assumptions of the satellite load and links evolution. Hence, the main driver of our solution is to devise a simple \emph{continual distributed learning} approach for each satellite to take decentralized decisions based on local observations and limited feedback. 

Building on our previous work in \emph{Q-routing}~\cite{soret2023q} and \gls{drl}~\cite{lozano2024multi}, this paper proposes a multi-agent continual learning framework for routing. This combines offline, global path calculations and online distributed model updates where each satellite participates in a data-driven aggregation for model alignment. Specifically, we use Q-learning, a model-free type of \gls{rl} that focuses on how agents interact with their environment and learn to improve their behavior by taking actions for which they expect a reward, i.e., by trial and error. 

\begin{figure}[t]
    \centering
    {\includegraphics[width=\textwidth]{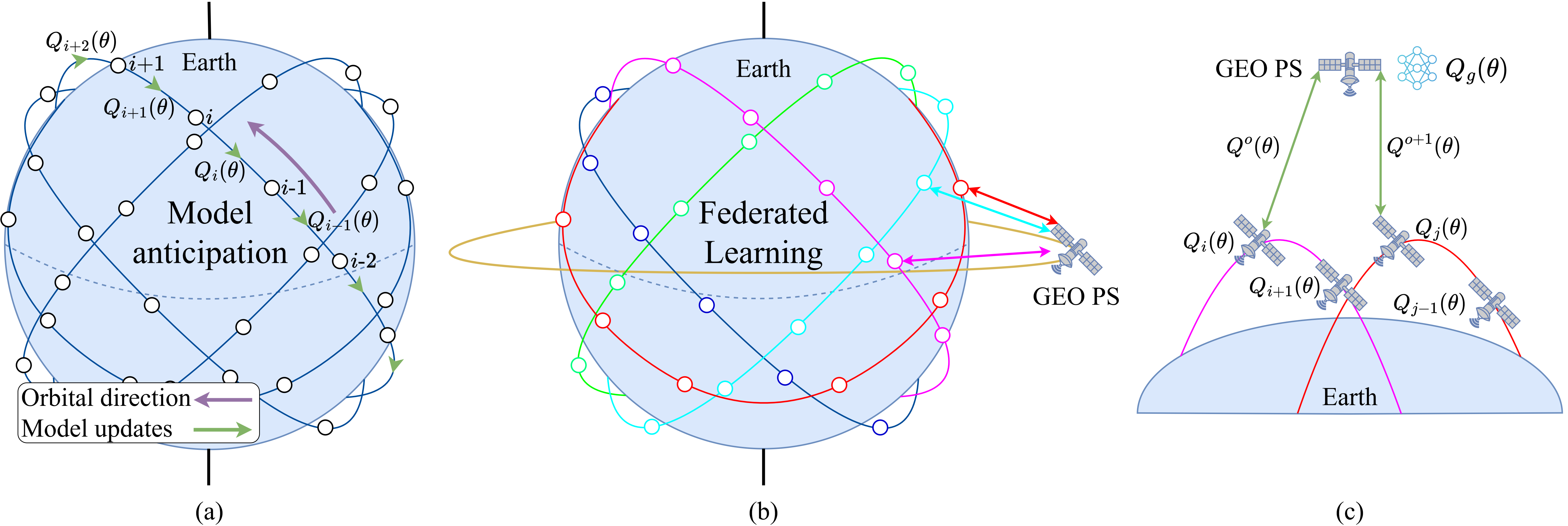}}
    \caption{Proposed solutions for continual decentralized learning in satellite constellations. (a) {Model anticipation: } Each agent $i$ transmits its Q-Network, $Q_i(\theta)$, to agent $i-1$ within the orbital plane and moving in the direction opposite to the orbital motion. (b) {Satellite Federated Learning (SFL): }Each orbital plane is a learning cluster formed by a set of satellite agents. The \gls{ps} at a \gls{geo} satellite aggregates the cluster models. (c) Detail of the cluster and network aggregation in SFL. The \gls{ps} receives each cluster model $Q^o(\theta)$ and aggregate into the global model $Q_g(\theta)$. 
    }
    \label{fig:HighLevel}\vspace{-0.5cm}
\end{figure}

For the routing problem, the information exchange between agents can greatly accelerate the learning speed as they get information about the congestion level of the network. Minimal multi-agent collaboration was exploited in \cite{lozano2024multi}, which proposes the use of \gls{drl} for coping with the large state space of a dense constellation and assumes a network-wide training in a centralized node. The satellites infer the next hop to forward an incoming packet, given the network topology, the destination, and the partial knowledge of the load and queues at the first neighbouring satellites. The pre-trained \gls{dnn} models are then uploaded on-board. However, these can diverge and face new challenges as time goes by and the environment evolves, for example, with some non-functional nodes or traffic variations. Therefore, this phase must be followed by an online update of the models, for which we propose two options, both enabling continual learning in the network:

\textbf{Model anticipation}: The first one leverages the predictability of the constellation movement. At time $t$, each satellite $i$ shares its Q-Network $Q_i(\theta)$ with the satellite behind within the orbit $i-1$, i.e., the one that will occupy a very close position in $t+\Delta t$ (see Fig.~\ref{fig:HighLevel}.a). We call this option \emph{model anticipation}. This case minimizes the signaling and overhead, and it is particularly interesting for networks with limited connectivity with the ground segment or the \gls{geo} layer. Moreover, the communication uses only the more stable intra-plane \gls{isl}. However, the alignment is not at the network level and therefore should be complemented in the long-term with a network-wise aggregation. 

\textbf{\gls{fl}}: The second option is \gls{fl} (Fig.~\ref{fig:HighLevel}.b), a decentralized collaborative approach for multiple partners to train locally and build a shared model with low bandwidth requirements. While there are previous works looking at the application of \gls{fl}\cite{matthiesen2023federated} in \gls{lsatc}, this paper breaks a new ground by using \gls{fdrl} for a concrete space application -- routing. Each satellite is continually training its pre-loaded model using the new available local data set, i.e., the new packets to be forwarded. Periodically, the agent Target Networks are first aggregated at the cluster level $Q^o(\theta)$, and then at the \gls{ps} which is the central orchestrator in charge of the aggregation process $Q_g(\theta)$ from outside the constellation (Fig.~\ref{fig:HighLevel}.c), located either in a \gls{gs} or a \gls{geo} satellite.

\subsection{Related work}


\subsubsection{Machine Learning:}
\gls{ml} and \gls{ai} have diverse applications in all layers of a space communications system, from the physical layer (e.g., Doppler shift estimation, channel modeling), to the data link layer (e.g., resource allocation, spectrum management, network slicing, anti jamming), and the upper layers (e.g., traffic forecasting, offloading, routing). The surveys in \cite{mahboob2024survey} and \cite{fourati2021artificial} provide recent overviews of this broad topic. 


The first learning area relevant for this work is \gls{rl}~\cite{sutton2018reinforcement}. \gls{rl} aims to train an agent to make adequate decisions in a system, called the environment, from pure experience. Common \gls{rl} problems include games~\cite{mnih2013playing} and autonomous and control systems. 
For problems with a discrete and small set of possible actions, \gls{ql} is a popular off-policy \gls{rl} method where the agent does not learn the policy itself but just keeps an updated table with the quality of each state-action pair. As the complexity of the environment increases, \gls{drl} is a particular type of \gls{rl} that exploits the power of multi layer deep neural networks for state representation and/or function approximations~\cite{arulkumaran2017drlsurvey}. Then, \cite{hasselt2010double} and~\cite{van2016deep} introduce \gls{dql} and its extension integrating \gls{dnn} with \gls{ddqn}, respectively. These methodologies address the overestimation bias in \gls{ql}, enhancing the stability and performance of learning in complex environments.

On the other side, \gls{fl} was first introduced by Google~\cite{mcmahan2017communication}, and it represents a paradigm shift towards decentralized machine learning, enabling collaborative and decentralized model training while preserving data privacy that can be integrated with \gls{rl}~\cite{zhu2023transfer}. Each agent has a model that can be pre-trained and improves it by updating it with local data. The local learnt information is summarized as a small focused update sent to a central \gls{ps}, which makes the aggregation either in a synchronous or asynchronous way \cite{li2021survey}. \gls{fl} is particularly convenient for the cases where the data is generated in the form of isolated islands \cite{yang2019federated}, which is the case in space \cite{matthiesen2023federated}. 
Recent works have explored the implementation of \gls{fl} within satellite clusters~\cite{razmi2023board, matthiesen2023federated}.

To cope with real-world dynamics, an intelligent system must have the ability to incrementally learn without forgetting knowledge acquired in the past, an ability termed \emph{continual learning} or lifelong learning~\cite{wang2024continual}\cite{khetarpal2020continual}. Specifically, in domain-incremental learning the algorithm must learn the same kind of task but in different contexts, e.g., due to a non-stationary stream of data. Despite the time-variant (and often non-stationary) nature of communication networks and traffic, this topic has received little attention so far. In the case of packet routing in \gls{lsatc}, as the satellite moves and the traffic distribution changes, the forwarding rules must be updated to reflect the changes in input-distribution. Recent papers have looked at continual learning in 
communication problems such as resource allocation~\cite{sun2022continual} and channel estimation~\cite{akrout2022continual}. Many other problems -- including routing -- remain unexplored. 


\subsubsection{Routing in \glspl{lsatc}:}
Routing in space networks has been investigated for many years, although there is lately a renewed interest in the topic due to the increasing complexity of the space missions, the need to move from proprietary to standardized \glspl{ntn} solutions, and the increasing intelligence of the network. 
There are two main strategies to address routing~\cite{xiaogang2016survey}, static and dynamic. The static routing schemes lies on the idea of addressing a changing network in a static manner by using \gls{vt} or \gls{vn} based methods. 
The \gls{vt}-based routing divides the \glspl{lsatc} period into time slices, treating the network topology as static within each slice~\cite{jia2018routing}~\cite{werner1997dynamic}\cite{roth2023analyzing}. In contrast, \gls{vn}-based routing overlays a virtual network on the Earth, mapping each virtual node to the nearest \emph{real} satellite~\cite{ekici2001distributed}.

In both approaches, \gls{vt} and \gls{vn}, the network is defined fixed in certain time periods or areas and both rely on centralized routing calculations. Therefore each satellite has to store routing tables according to time and position, making it challenging and load-expensive to update all the network simultaneously due to the needed signaling.~\cite{roth2023analyzing} proposes that the signaling needed to update the tables can be reduced assuming that the network changes are predictable, however, link failures and congestion due to traffic load changes are not considered.

Dynamic routing was first studied in the context of ad-hoc terrestrial networks, specifically \gls{dsdv}~\cite{perkins1994highly} and \gls{aodv} routing protocols~\cite{perkins2003ad} use proactive and reactive principles, respectively. In general, most current dynamic routing strategies rely on data packets flooding and broadcasting, consuming too much satellite computational resource and network bandwidth~\cite{li2019temporal}.

Alternatively, \gls{rl}-based routing algorithms allow to have a dynamic routing strategy where each agent -- the satellite -- takes autonomous forwarding decisions based on partial observations. Specifically, \gls{qr} is a distributed and autonomous packet routing algorithm based on \gls{ql} and first proposed in~\cite{boyan1993packet}. Building on this work, ~\cite{huang2023reinforcement} proposed a distributed \gls{ql}-based routing algorithm for \glspl{lsatc} where each node is a different agent and the routing strategy is updated in a distributed manner. This is similar to our initial work in\cite{soret2023q}, but in~\cite{huang2023reinforcement} the Q-tables lack of positioning and congestion information, plus they have to be re-trained as the satellites move. This lack of flexibility is solved with the introduction  of \glspl{dnn}. 
\cite{liu2020drl} proposes a centralized \gls{drl}-based routing protocols in satellite mega-constellations, demonstrating the potential of learning-based, energy-aware routing protocols. However, this model relies on a network controller, which is an agent with real-time full knowledge of the network. 
\cite{guoliang2022marouting} presented a fully distributed algorithm aiming to minimize the average latency, although it does not explicitly address the ground segment, focusing solely in the space segment. The authors in~\cite{lyu2024marouting} formulate a constrained-\gls{ma-drl} problem for the packet routing in an integrated satellite-terrestrial network, aiming at meeting delay and energy constraints.  

The existing literature lacks a full \gls{ma-drl} solution for on-board routing in a \gls{lsatc} with non-stationary data, considering the three main sources of uncertainty -- the links, the buffers, and the traffic -- and the limited knowledge at each satellite. In this paper, we build upon our previous work in~\cite{soret2023q}, where we proposed a decentralized \emph{Q-routing} algorithm that was then extended into a \gls{ma-drl} framework~\cite{lozano2024multi}. We focus now in the lifelong learning of the routing by proposing a \gls{ma-drl} for in-orbit learning based on \gls{fl}, enabling incremental learning and dynamic knowledge sharing among satellites.

\subsection{Contributions and organization}
The main contributions of this paper are:
\begin{itemize}
    \item We model the general scenario of routing over a \gls{lsatc} considering data packets from a ground source to a ground destination through the constellation, the space and ground time-varying links, and the communication buffers.  
    \item We formulate a \gls{pomdp}~\cite{oliehoek2016concise} for taking routing decisions in \glspl{lsatc} with low E2E delay. Each satellite acts as an agent of the \gls{ma-drl} framework, with the actions being the next hop (agent) to forward the packet. Minimal feedback from nearby agents is designed to cope with the limited observability.  
    \item We propose a continual learning framework to evolve with the environment. The initial offline phase has decentralized decisions and a global training, while the online, on-board phase relies on pre-trained \glspl{dnn} and has a long time span for the lifetime of satellites. During the online, a short-term model anticipation exploits the predictability of the satellite movements, and a \gls{fl} performs model aggregation for long-term alignment of the models. 
    \item We verify the potential of our proposal with extensive simulations of the offline, online, and continual learning with state-of-the-art \glspl{lsatc}. 
\end{itemize}

The rest of the paper is organized as follows. Section~\ref{sec:systemmodel} introduces the communication system model. The learning framework is presented in \ref{sec:Learning_framework}, with the offline and online phases described in Sections~\ref{sec:Offline_learning} and ~\ref{sec:Online_learning}, respectively. The proposal is evaluated in Section~\ref{sec:results} and concluding remarks are given in Section~\ref{sec:conclusions}. 

\section{System Model} \label{sec:systemmodel}
Table~\ref{tab:system_params} has a list of the notation used in this paper. Although the \gls{lsatc} is moving and the related communication parameters change with time, we omit the time index for notation simplicity.

\begin{table}[t]
\centering
\caption{Notation of variables and corresponding description}
\renewcommand{\arraystretch}{1.2}
\begin{tabularx}{\columnwidth}{@{}llX@{}}
\toprule
\multicolumn{2}{@{}l}{Notation} & Description \\\midrule
\multicolumn{3}{@{}l}{\textbf{Graph}}\\ 
&$\mathcal{G}(\mathcal{N},\mathcal{E})$ & Multi-partite graph with nodes $\mathcal{N}$ and edges $\mathcal{E}$ \\  
&$\mathcal{N} = \mathbb{S} \cup \mathbb{G}$& The set of satellites $\mathbb{S}$ and the set of gateways $\mathbb{G}$ are the two subsets of nodes in $\mathcal{G}$ \\
&$\mathcal{E} = \mathcal{E}_S \cup \mathcal{E}_G$& The set of edges as the sum of the edges of the space and ground segment \\
&$\mathcal{E}_S = {\bigcup_{i\in \mathcal{S}}\mathcal{E}_{i_S}}$& The set of satellite edges as the {union} of the edges of each satellite $i$ \\
&$||ij||$ & Slant range between nodes $i$ and $j$ in the graph [m]\\
&$\mathcal{P}$ & Set of packets\\
&$h$ & Altitude of deployment of the satellites $[\mathrm{m}]$\\
&$T_o$ & Orbital period of the orbital plane $o$ $[\mathrm{s}]$\\
\multicolumn{3}{@{}l}{\textbf{Communications}}\\ 
&$R(i, j)$ & Data rate for transmission from node $i$ to node $j$ $[\mathrm{bps}]$\\
&$D(i, j)$ & Total delay for transmission from node $i$ to node $j$ $[\mathrm{s}]$\\
&$\text{SNR}_{\min}(\rho)$ & Minimum Signal to Noise Ratio for reliable communication with spectral efficiency $\rho$ \\
&$\lambda^g_{UL}$ $(\lambda^g_{DL})$ & Uplink (donwlink) data generation at gateway $g$\\
&$Q^{\max}$ & Size of the communication queues at the satellites $[\mathrm{b}]$\\
\multicolumn{3}{@{}l}{\textbf{Learning}}\\
&$S_t^i$ & State at time $t$ observed by agent $i$\\
&$a^i_t$ & Action at time $t$ by agent $i$\\
&$r^i_t = r_q + r_d + r^\star$ & Reward at time $t$ by agent $i$ as the sum of the reward related to the queue occupancy $r_q$, getting closer to the destination $r_d$, and extra penalties and rewards $r^\star$\\
&$Q_i(S_t^i, a^i_t; \theta)$ & Q-Network of \gls{drl} agent $i$ with weights $\theta$ at time $t$ \\
&$Q_g(\theta)$ & Global Q-Network in the offline training\\
&$Q^o(\theta)$ & Aggregated cluster Q-Network in orbital plane $o$\\
     \bottomrule
\vspace{-1em}
\end{tabularx}

\label{tab:system_params}
\end{table}

\gls{e2e} routing takes place over a \gls{lsatc}
to connect two points (gateways) on Earth's surface, integrating both space and ground segments into the communication network. Mathematically, this is abstracted as a dynamic graph $\mathcal{G}(\mathcal{N}, \mathcal{E})$ formed by the set of nodes $\mathcal{N}$ and the set of edges $\mathcal{E}$. 

\noindent \textbf{Space segment.}
The \gls{lsatc} consists of $N$ satellites evenly distributed across $O$ orbital planes 
, forming a finite set of satellite nodes, $\mathbb{S}$, and a set of edges, $\mathcal{E}_S$, representing the transmission links between them. We remark that $i \in\{1,2,\dotsc, N\}$ indexes the satellites and, since each satellite is a learning agent in our model, it is later used to index agents in the learning framework.
Each orbital plane $o\in\{1,2,\dotsc,O\}$, typically approximated as a circular orbit, is deployed at a given altitude $h_o$~km above the Earth's surface, at a given longitude $\epsilon_o$~radians, has a given inclination $\delta$, and consists of $N_o=N/O$ evenly-spaced satellites. 
In the orbit, each satellite $i$ follows a trajectory around Earth with orbital period $T_o = 2\pi \sqrt{a_i^3/\mu}$ where $a_i$ is the semi-major axis and $\mu=3.98 \text{x} 10^{14} \text{m}^3/\text{s}^2$ is the geocentric gravitational constant. Each satellite is equipped with one antenna for ground-to-satellite communication and four antennas for inter-satellite communication. Two of the latter antennas are located at both sides of the roll axis (i.e., front and back of the satellite) and are used to communicate with immediate neighbors within the same orbital plane (intra-plane \gls{isl}). The other two antennas are located at both sides of the pitch axis and are used to communicate with satellites in different orbital planes through inter-plane \gls{isl}\cite{Leyva-Mayorga2021}. We denote $\mathcal{E}_{i_S}$ the set of feasible {inter-satellite} edges of satellite $i$, i.e., the \gls{isl} that are currently available for communication. $\mathcal{E}_{i_S}$ has a maximum size of four: $|\mathcal{E}_{i_S}| \leq 4$. $\mathcal{E}_S$ is the set of edges of the space segment, i.e., $\mathcal{E}_S = \bigcup_{i\in \mathbb{S}} \mathcal{E}_{i_S}$. Choosing the best set of feasible \gls{isl} links {$\mathcal{E}_{i_S}$} for each satellite $i$ at each time instant is a dynamic matching problem. For a constellation with a uniform grid a \emph{greedy approach} is applied as follows. Each satellite establishes intra-plane \gls{isl} with its immediate upper and lower neighbors within the orbital plane. For inter-plane connectivity, a satellite reaches out to the closest counterparts in adjacent planes to the east and west, prioritizing proximity to minimize latency and optimize data rates. This iterative process has to consider the bidirectionality of the links. Otherwise, for general constellations, we apply the algorithms proposed in \cite{Leyva-Mayorga2021}.


\begin{figure}[t]
\centering
\begin{subfigure}{0.48\textwidth}
    \centering
    \includegraphics[width=\linewidth]{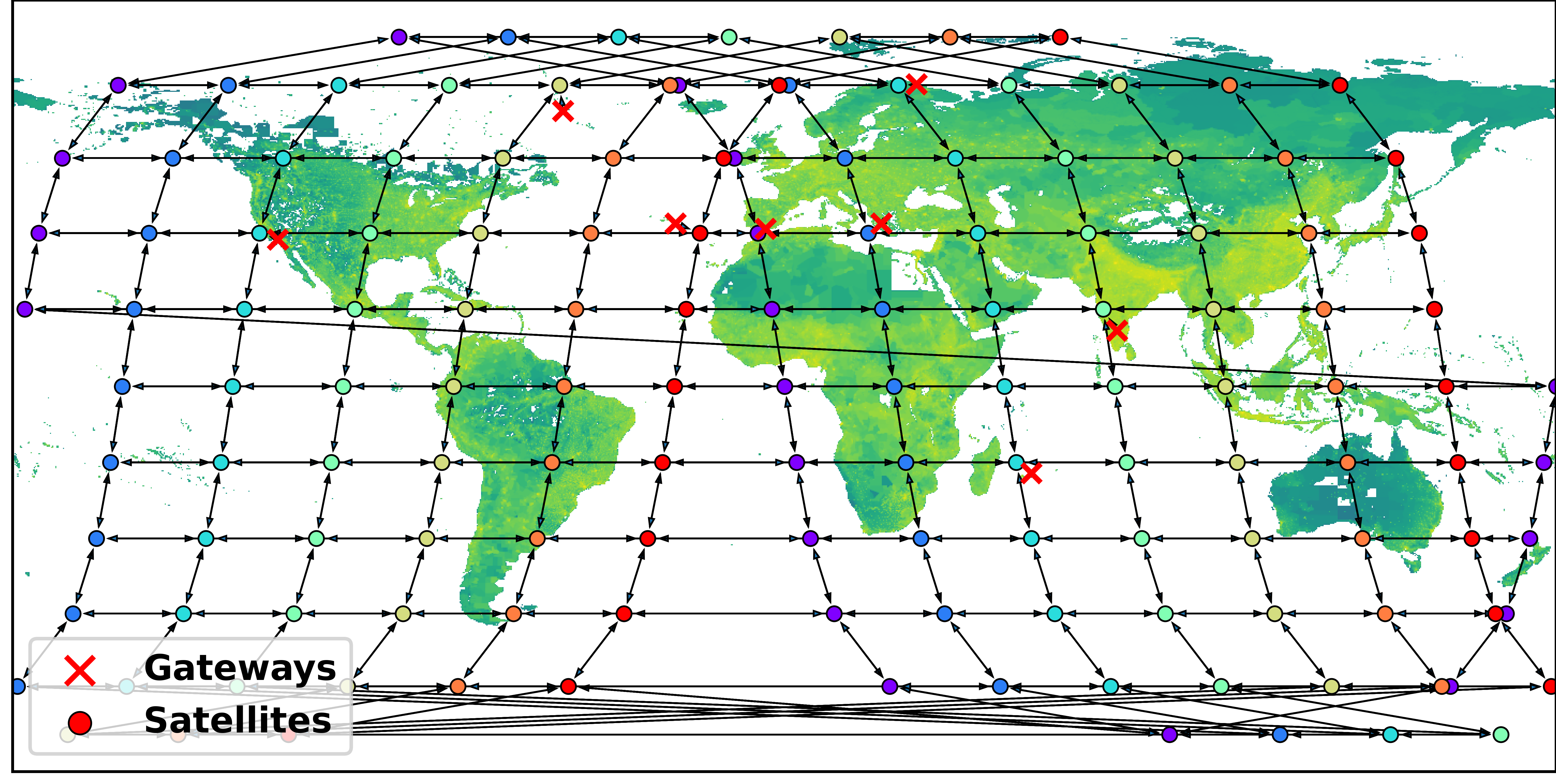}
    \caption{Kepler constellation}
    \label{fig:ISLs_Kepler}
\end{subfigure}%
\hfill
\begin{subfigure}{0.48\textwidth}
    \centering
    \includegraphics[width=\linewidth]{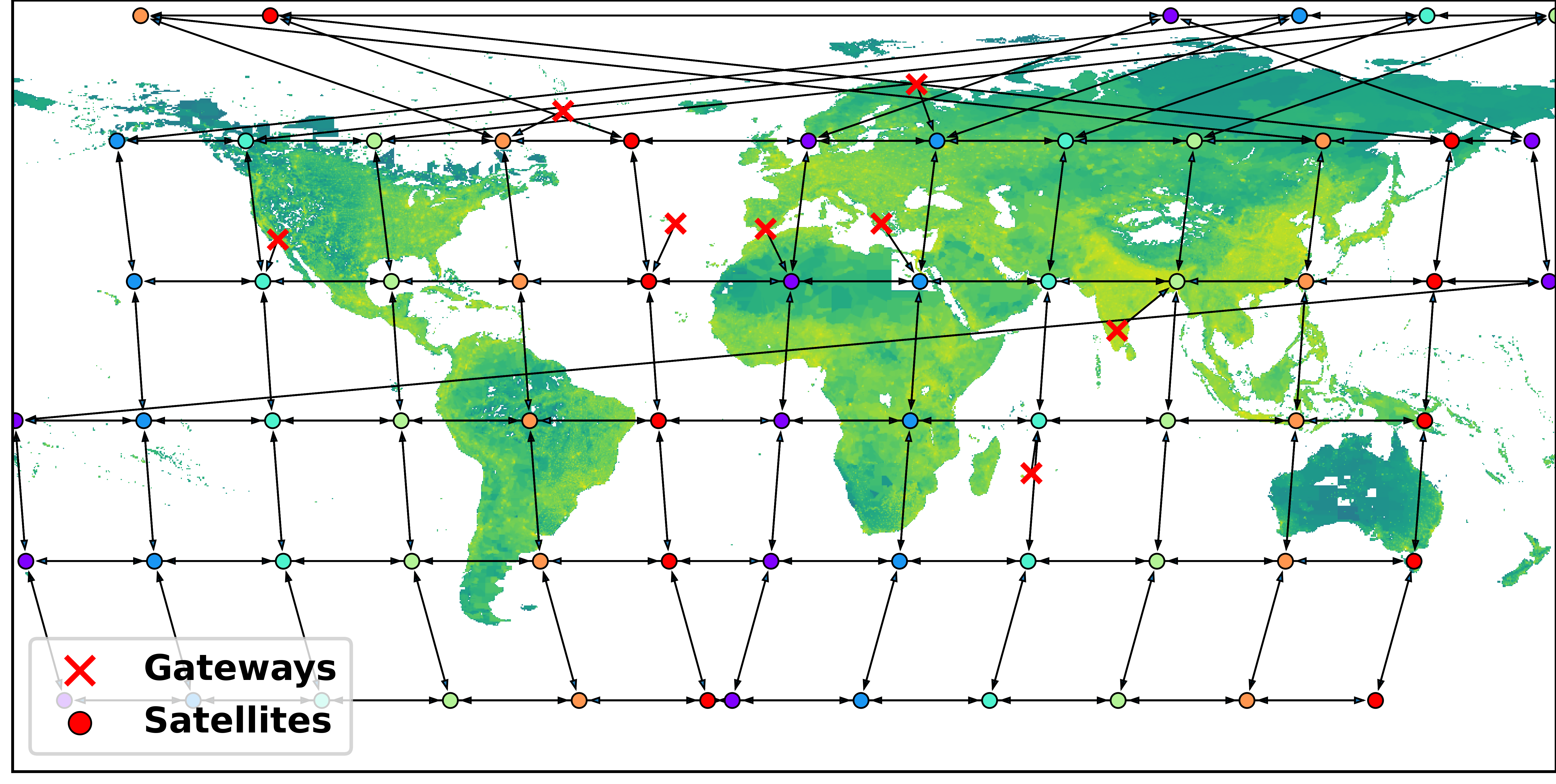}
    \caption{Iridium NEXT constellation}
    \label{fig:ISLs_Iridium}
\end{subfigure}
\hfill
\begin{subfigure}{0.48\textwidth}
    \centering
    \includegraphics[width=\linewidth]{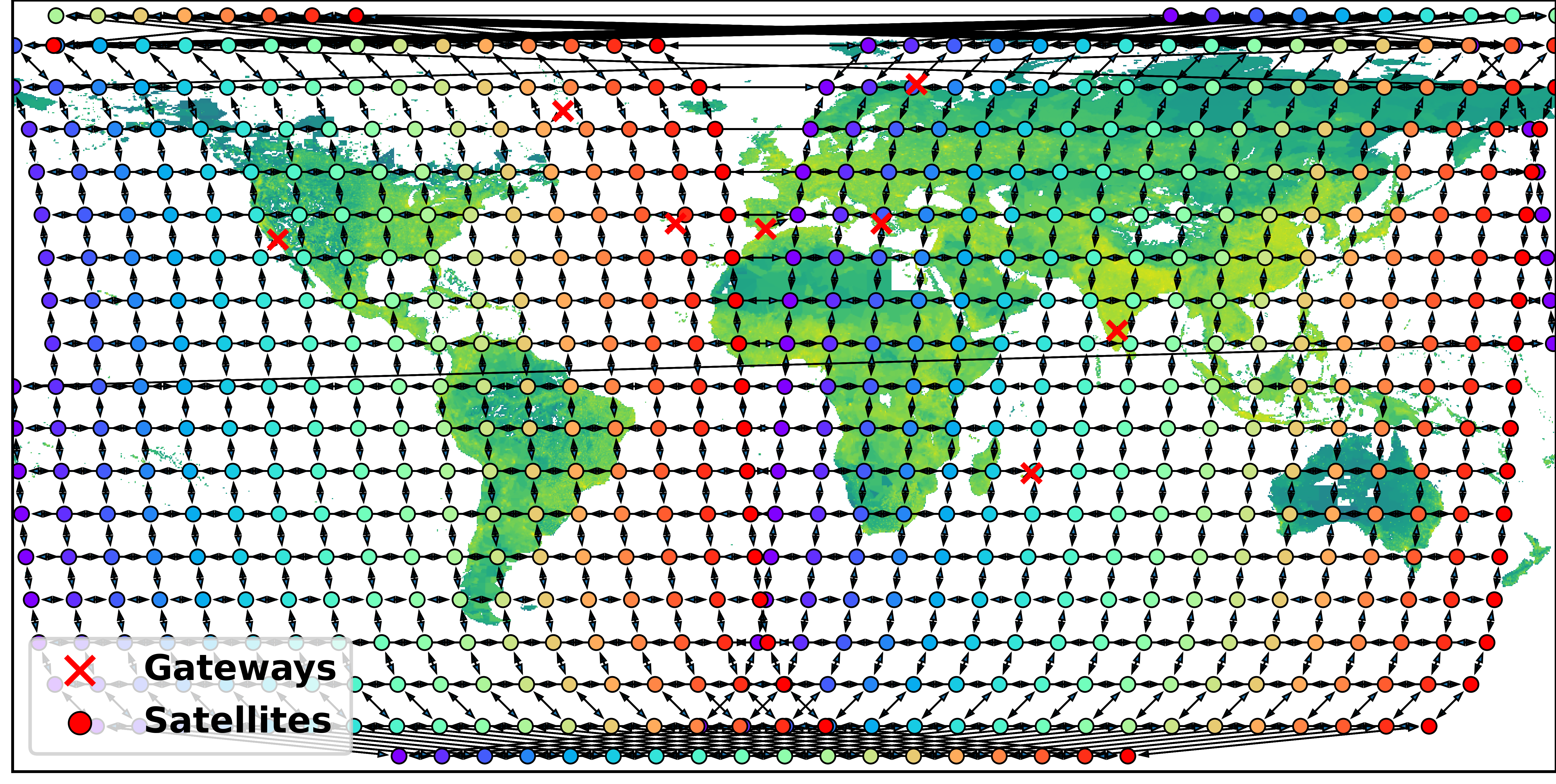}
    \caption{OneWeb constellation}
    \label{fig:ISLs_OneWeb}
\end{subfigure}
\hfill
\begin{subfigure}{0.48\textwidth}
    \centering
    \includegraphics[width=\linewidth]{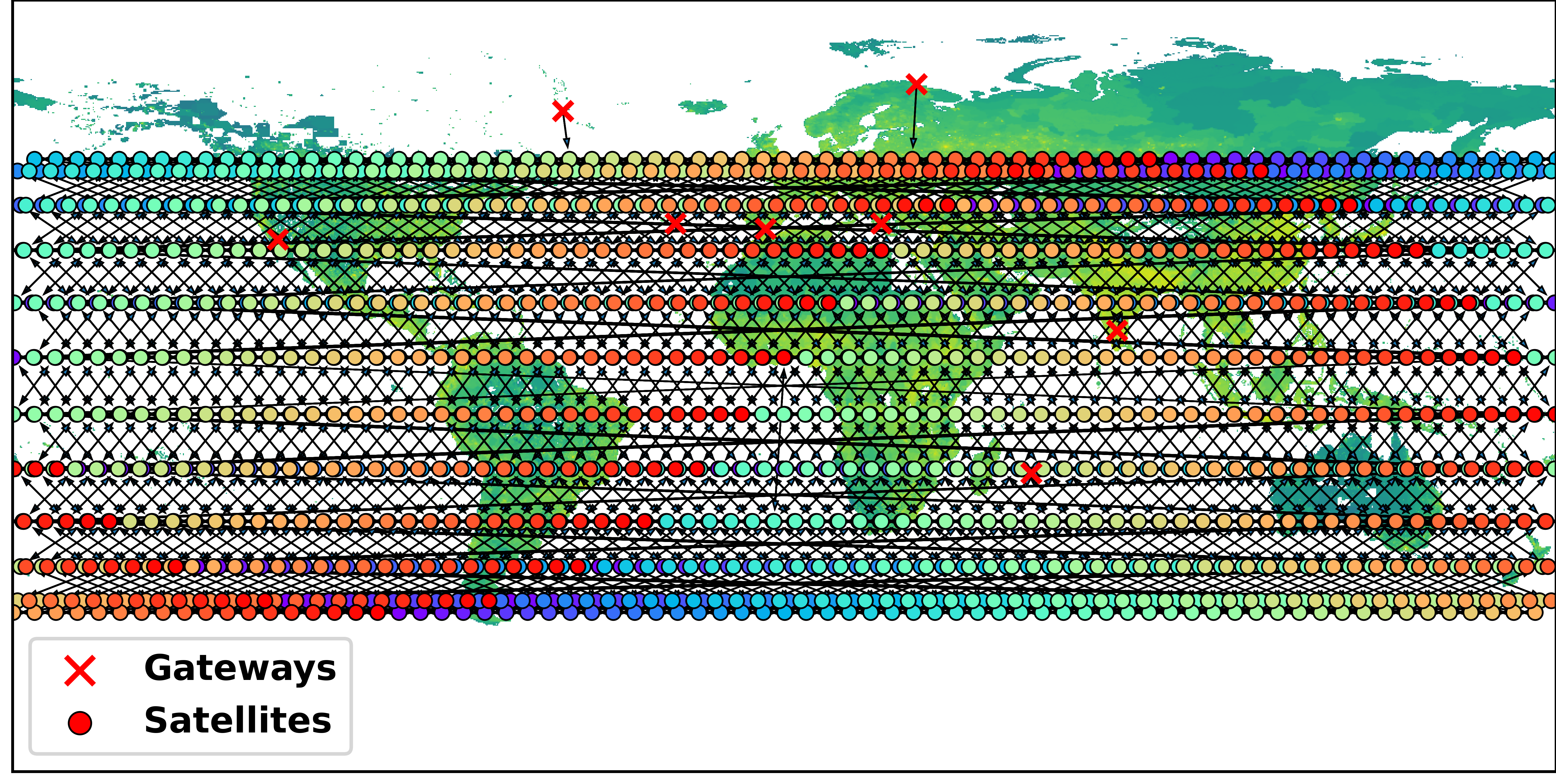}
    \caption{Starlink constellation}
    \label{fig:ISLs_Starlink}
\end{subfigure}
\caption{Representation of the four tested constellations and their corresponding \glspl{isl} established following the \emph{Greedy matching} over the population maps~\cite{CIESIN}. Each colour is a different orbital plane.} 
\label{fig:ISLs}\vspace{-0.6cm}
\end{figure}


\noindent \textbf{Ground segment.} 
The ground segment includes a set of gateways $\mathbb{G}$ distributed across the Earth, 
in locations denoted as $\left(g_\text{lat},g_\text{lon}\right)$ for $g\in\mathbb{G}$, where $g_\text{lat}$ and $g_\text{lon}$ are the latitude and longitude, respectively. The mobile devices at ground level communicate with the constellation through these gateways, which collect and aggregate the data from nearby users into relatively large packets that are transmitted to their linked satellite (a.k.a. service link). The number of mobile users is extracted from a population map~\cite{CIESIN}, which divides the Earth into a set of regions $\mathcal{R}$, evenly spaced by $0^\circ 15^\prime$ in both latitude and longitude. For each region $r\in\mathcal{R}$, the map indicates its mid-point $(r_\text{lat},r_\text{lon})$ and the number of users in the region $N_r$. 

Each gateway keeps one \gls{gsl} with their closest satellite at all times. These \glspl{gsl} constitute the set of edges $\mathcal{E}_G$, and $\mathcal{E}_{i_G}$ is the \gls{gsl} between a gateway and its closest satellite if it is within a visible range.
Therefore, there are three cases for the link $ij$ between node $i$ and node $j$: (1) $i\in\mathbb{G}$ and $j\in\mathbb{S}$ corresponds to the uplink from a gateway to a satellite, and $ij \in \mathcal{E}_G$; (2) $i\in\mathbb{S}$ and $j\in\mathbb{G}$ corresponds to the downlink from a satellite to a gateway, and $ij \in \mathcal{E}_G$; (3) otherwise, both $i$ and $j$ are satellites and we have an \gls{isl}, and $ij \in \mathcal{E}_{i_S} \subset \mathcal{E}_S$. The slant range between nodes $i$ and $j$ is denoted as $||ij||$. In a nutshell, each satellite and each gateway is a different node in $\mathcal{G}$, i.e, $\mathcal{N} = \mathbb{S} \cup \mathbb{G}$, and each \gls{isl} and each \gls{gsl} is a different edge in $\mathcal{G}$, i.e, $\mathcal{E} = \mathcal{E}_S \cup \mathcal{E}_G$.


\noindent \textbf{Data rate.}
The communication data rate between nodes $i$ and $j$, for both \gls{isl} and \gls{gsl}, is selected for the feasible links as the highest modulation and coding scheme that ensures reliable communication given the current \gls{snr}, and zero otherwise. To operate with realistic data rates, we denote $\left\{\rho\right\}$ bits/s/Hz as the set of spectral efficiencies achievable in the DVB-S2 technology~\cite{dvb_s2}. Moreover, let $\text{SNR}_\text{min}(\rho)$ represent the minimum \gls{snr} required for reliable communication with $\rho$. Assuming free-space path loss, the equation for calculating the data rate for transmission from $i$ to $j$ is given by:

\begin{equation} \label{eq:rate}
R(i,j) = W\max\left\{\rho : \frac{P_r(i,j)}{k_BT_S\ilm{W}} \geq \text{SNR}_\text{min}(\rho)\right\},
\end{equation}

\noindent where $W$ is the bandwidth, $P_r(i,j)$ denotes the power received by node $j$ from $i$,  including the antenna gains and free-space path loss, $k_B$ is the Boltzmann's constant, and $T_S$ is the system noise temperature.


\noindent \textbf{Traffic generation.}
We consider a scenario with realistic packet generation, queuing, and transmission, where each gateway transmits an equal amount of data among the rest of the gateways through the \gls{lsatc}, which then distribute the data among the users connected to them. 
Let $\lambda_\text{UL}^{(g)}$ be the uplink data generation (i.e., arrival) rate at gateway $g$ and {$\lambda^*$  be the maximum supported traffic load in the network, calculated from the uplink and downlink \gls{gsl} rates. Next, we define the total traffic load as  $\ell=\sum_{g\in\mathbb{G}}\lambda_\text{UL}^{(g)}/\lambda^*$. Then, the amount of data transmitted to a gateway is the aggregation of contributions from the other gateways, assuming that the uplink traffic of each gateway is split equally across the other $|\mathbb{G}|-1$ gateways:
\begin{equation}    
\lambda_\text{DL}^{(g)}=\sum_{i\in\mathbb{G}\setminus \{g\}} \frac{\lambda_\text{UL}^{(i)}}{|\mathbb{G}|-1}=\frac{\ell\lambda^*-\lambda_\text{UL}^{(g)}}{|\mathbb{G}|-1}\leq R(i,g),
\end{equation}
where $R(i,g)$ is the downlink data rate for gateway $g$.}
Data is generated at the gateways following a Poisson distribution with rate $\lambda_\text{UL}^{(g)}$ and a block of $B$\,bits is sent when a source gateway has $B$ bits with the same destination gateway.  The set of packets transiting the network is chronologically sorted in $\mathcal{P}$.


\noindent \textbf{Routing.}
The routing algorithm at each satellite $i$ aims at relaying each received packet $p(d)$ towards the destination $d$. If $i$ is directly connected to $d$, $p(d)$ is forwarded there, otherwise the routing decision is taken in a decentralized manner at each node. Each satellite is equipped with a transmission buffer with a maximum capacity of $Q^{\max}$, operating under a \gls{fifo} strategy. If the buffer is not empty, the satellite takes the \gls{hol} packet and delivers it to one of its linked nodes. Any packet arriving at a full buffer will be dropped. The set of nodes visited by packet $p$ is denoted by $\mathcal{P}_p$.


\noindent \textbf{Latency.}
The one-hop latency to transmit a packet $p$ of length $B$~bits from $i$ to $j$ is computed as follows. 
First, the \textit{queue time} at the transmission queue $t_q(i)$ is the time elapsed since the packet is ready to be transmitted until the beginning of its transmission. The transmission buffer at a satellite $i$ has a length $q_i$ and follows a \gls{fifo} policy. Although we consider a transmission from $i$ to $j$, the rest of packets in the queue may have different destinations and therefore use different links from the set $\mathcal{E}_{i_S}$ and $\mathcal{E}_{i_G}$, with different rates $R(i,\cdot)$. The queue time is hence given by $t_q(i) = \frac{q_i \, B}{R(i,\cdot)}$ where $R(i,\cdot)$ is calculated from (\ref{eq:rate}) and depends on the link used for the transmission of each en-queued data packet ahead of $p$. Second, the \textit{transmission time}, which is the time it takes to transmit $B$~bits at $R(i,j)$~bps. Third, the \textit{propagation time}, which is the time it takes for the electromagnetic radiation to travel the distance $||ij||$ from $i$ to $j$. The total one-hop delay for the link from $i$ to $j$ is written

\begin{equation} \label{eq:latency_metric}
    D(i,j) = { t_q(i) }+ { \dfrac{ B }{ R(i,j) } }+\dfrac{||ij||}{c}.
\end{equation}


This latency model accounts for varying traffic loads, where propagation time dominates in a non-congested satellite constellation, but queue time quickly escalates under high traffic conditions~\cite{Rabjerg2021}.


\section{Learning framework} \label{sec:Learning_framework}

\begin{figure}[t]
    \centering
    {\includegraphics[width=0.48\textwidth]{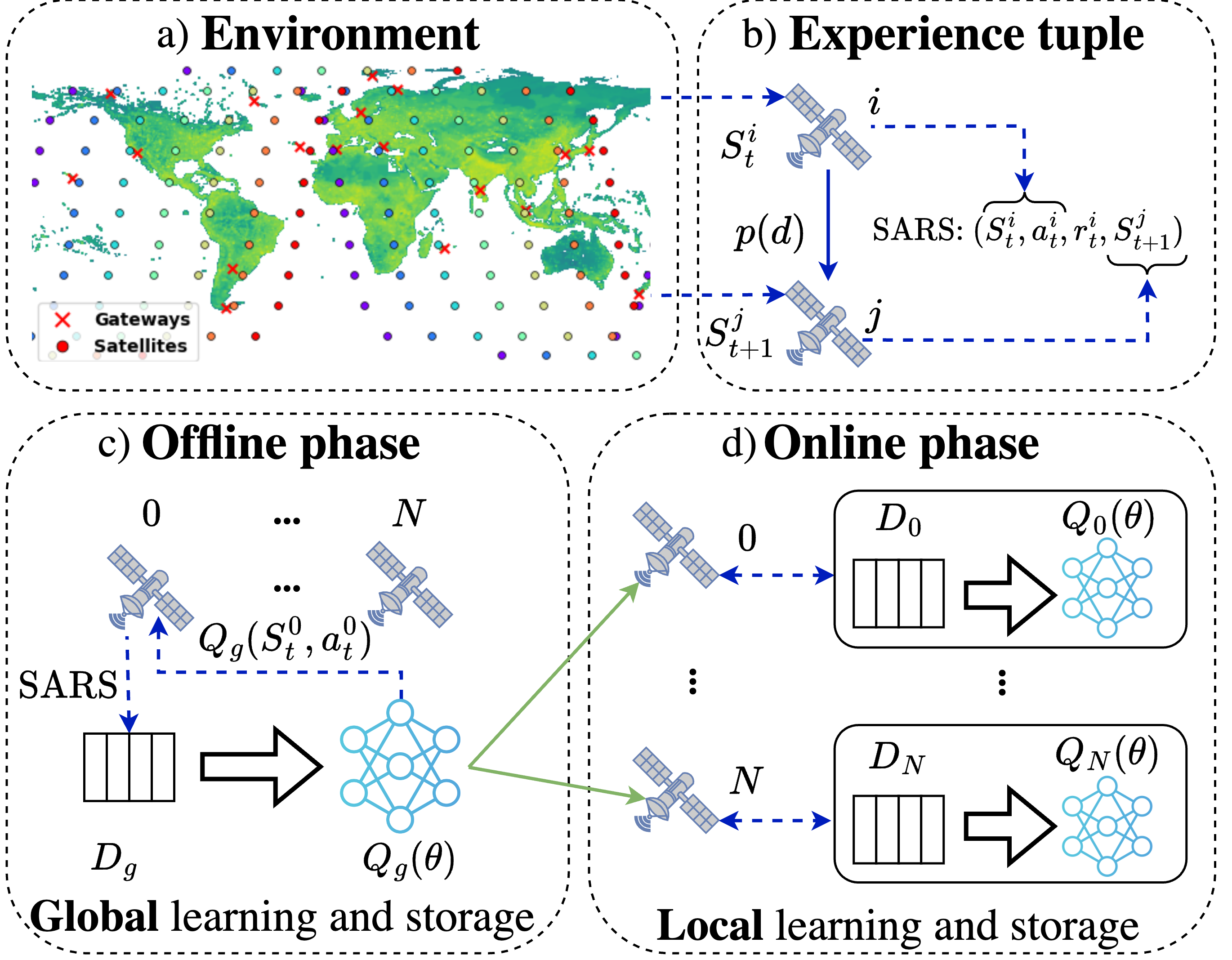}}
    \caption{Proposed learning framework, where: a) presents the environment {with a Kepler constellation}, including the network topology; b) represents a network-agent interaction needed to build the tuple experience ($\text{SARS}$); c) represents the offline phase; and d) represents the online phase.}
    \label{fig:Phases}\vspace{-0.5cm}
\end{figure}

The proposed learning framework is illustrated in Figure~\ref{fig:Phases}. {Each satellite is a different agent that makes routing decisions around the globe in a decentralized manner relying upon local observations.}
\ilm{Agents operate in a time-varying environment {(Fig.~\ref{fig:Phases}.a)} where the local experience gathered at each time step is stored as a tuple {(Fig.~\ref{fig:Phases}.b)}.} The \gls{drl} process is structured into two phases. During the offline phase (Fig.~\ref{fig:Phases}.c), \ilm{a high exploration rate is employed in the routing decisions, which favours the long-term learning of the system over the short-term.} 
{A global model gets the observed state and determines the value of each state-action pair. Next, the experience tuples of each agent are stored globally, such the global model} can learn from all them within a simulation.
Subsequently, in the online phase (Fig.~\ref{fig:Phases}.d), the global model pre-trained in the offline phase is uploaded on-board of each agent so now each one has its own local model and buffer. This phase emphasizes exploitation and now each agent keeps training its own model with its own local experiences.


The \gls{lsatc} represents a networked multi-agent problem where agents are partially connected and interact uniquely with nearby agents. In this \gls{ma-drl} we aim at a fully-decentralized solution where each satellite acts as an independent agent that learns its own routing policy. Since each agent has only local information of the observed environment based on the local network state and the restricted communication with their one-hop neighbouring satellites, the decision-making problem is based on a partially observable state.

The problem is formulated as a \gls{pomdp}~\cite{oliehoek2016concise} with a 4-tuple $(\mathcal{S}, \mathcal{A}, P(s,a), \mathcal{R}(s,a))$, where $\mathcal{S}$ is the state space, $\mathcal{A}$ is the action space, $P(s,a)$ represents the probabilities of taking action $a$ in state $s$, and $\mathcal{R}(s,a)$ denotes the corresponding reward received. The time is discrete and starts at zero. For each time step $t$, an agent $i$ observes a state {$S^i_t$} within $\mathcal{S}$, selects an action $a^i_t$ within $\mathcal{A}$, gets the outcome reward $r^i_t$ from the environment, and observes the new state {$S^i_{t+1}$} after the state-action pair {$(S^i_t, a^i_t)$.}

 

A space environment is often constrained by the computing, communication and energy resources. It is therefore important to limit the complexity of the \gls{dnn} without jeopardizing the performance, for which we carefully design the state space as follows. 
The state space of agent $i$ is defined as $S^i : \{\mathcal{L}_i, \mathcal{N}_i \}$, where $\mathcal{L}_i$ is the local information at $i$, and $\mathcal{N}_i$ is the information shared by the two intra-plane and the two inter-plane neighbours satellites to $i$. 
Specifically, $\mathcal{L}_i$ is the information of the packet destination $d$ extracted from the packet header, the link connectivity $\mathcal{E}_{i_S}$, the current coordinates, and the neighbors' coordinates. $\mathcal{N}_i$ contains information about the congestion level of the four buffers of the first neighbours,  parameterized through $C_{j,k}$  as:


\begin{equation}
\label{eq:queue_val}
    C_{j, k} = \max\left( C^*, \left\lfloor C^* \cdot \frac{\log_{10}(q_{t} + 1)}{\log_{10}(q^{\max})} \right\rfloor \right)
\end{equation}

\noindent where $C_{j, k}$ is the encoded congestion value for the neighbor $j$'s queue $k$. $C^*$ is the maximum value of the encoded congestion, $q^{\max}$ is the queue size, and $q_{t}$ is the queue length at time $t$. Each of the four neighbors transmits the congestion level of its four queues. Therefore, $\mathcal{N}_i$ is a vector of 16 values which makes it manageable for the \gls{dnn}.

If a link to a neighbour is unavailable, then the congestion levels of all its queue lengths are set to infinite and its coordinates to {$0$}. If just a neighbours link is unavailable, then that queue length is set to infinity. Therefore, any observed state $S^i_t$ by an agent $i$ is defined as a vector of 28 fields: 16 for the neighbors' congestion queue levels, 8 for the neighbour coordinates, 2 for the current coordinates, and 2 for the packet destination $d$ connected satellite coordinates.

The satellite's absolute latitude and longitude are normalized by a granularity factor $\sigma$ to fall within the range $[0^{\circ}, 180^{\circ}/\sigma]$ and $[0^{\circ}, 360^{\circ}/\sigma]$ respectively. Meanwhile, the coordinates of neighboring and destination satellites are encoded relative to the current satellite's position, smoothly handling the discontinuity at the $180^{\circ}$ latitude boundary. This relative positioning yields a normalized range of $[-90^{\circ}/\sigma, 90^{\circ}/\sigma]$ for latitude and $[-180^{\circ}/\sigma, 180^{\circ}/\sigma]$ for longitude after applying the necessary shifts and scaling factors. 
Consequently, all fields in $\mathcal{S}$ are scaled uniformly, with congestion levels and positional values maintaining a consistent range for the \gls{dnn} inputs. 
Our approach enhances the logical location representation method described by~\cite{ekici2001distributed} by incorporating dynamic queue behavior, which their model did not consider, and incorporating the exact satellite coordinates scaled, which gives real-time positioning precision.

Regarding the action, $a_t^i$ is the next hop $j$ selected by agent $i$ from the set $\mathcal{E}_{i_S}$, i.e., one of the neighbouring satellites. An agent can also be connected to a gateway, however, if this is the destination of the packet this will be directly forwarded there without making any decision. Therefore, $\mathcal{A}$ has 4 possible {options}. 


Finally, the reward $r^i_t$ for the state-action pair $(S^i_t, a^i_t)$ at satellite $i$ has three terms:

\begin{equation}
\label{eq:reward}
r^i_t = w_1 \cdot r_q + {w_2 \cdot r_r + w_3 \cdot r^\star},
\end{equation}

\noindent where $w_1, w_2, w_3$ are weighting parameters, and the terms are defined next.

The first term is related to the time spent on the receiving agent $j$ queue, specifically 
\begin{equation}
    r_q = 1 - 10^{t_q(i)}
\end{equation}

\noindent being $t_q(i)$ the time $p$ has spent at the queue of the receiving agent. 

The second one captures how good is the action in getting closer to the destination:
\begin{equation}
r_r = \frac{||id||-||jd| - \frac{||ij||}{w_4}}{d^{\max}}
\end{equation}
\noindent where the slant range reduction is $||id||-||jd||$, and $||ij||$ is the distance traveled by a packet $p$ after being forwarded are included. The term $w_4$ is a weighting factor ensuring a positive numerator, as there is often a mismatch between slant ranges and physical distances following the orbital paths. Finally, this is normalized by $d^{\max}$, which is the maximum \gls{isl} distance between satellites in the constellation.

The last term is written
\begin{equation}
    r^\star = \mathds{1}\left(a^i_t = j \wedge j\in \mathcal{P}_p\right)\cdot r_l + \mathds{1}\left(d \in \mathcal{E}_G \right)\cdot r_d + \mathds{1}\left(a^i_t \notin \mathcal{E}_{i_S}\right)\cdot r_u, 
\end{equation}

\noindent where $\mathds{1}(\cdot)$ is the indicator function and $\wedge$ is the logical conjunction. This term $r^\star$ includes extra penalties and rewards as follows: 
$r_l$ is a penalty applied when a packet revisits a node {in order to avoid looping}; $r_d$ is a reward added when the agent has a \gls{gsl} to the destination gateway; $r_u$ is a penalty applied to the action that aims at forwarding the packet through an unavailable link.

In this manner every agent is being rewarded for sending the packets to other non-congested agents that are closer to the packet destination, accumulating less distance travelled {and less time spent at queues} as possible and ensuring that extreme cases are penalized or rewarded accordingly. 
This reward function aims at minimizing the queue time and the distance, as the main contributors of the end to end (E2E) delay are the propagation time and the queuing time when the network is uncongested and congested, respectively~\cite{Rabjerg2021}. However, other parameters can be optimized. For instance, an agent can receive an extra reward when forwarding a packet to an intra-plane neighbour agent, favouring this behaviour.

In conventional Q-learning, the agent seeks the optimal policy, which is the set of state-action pairs maximizing the long-term accumulated reward. The state-action Q-value, $Q_i(S^i_t, a^i_t)$, represents the estimated value of taking action $a^i_t$ in state $S^i_t$ at time $t$, i.e, the expected future cumulative rewards. The agent updates the Q-values through exploration by learning from the \gls{td} error, which quantifies the discrepancy between the current Q-value estimation and the new estimate after the action $a^i_t$ is taken based on the received reward $r^i_t$ plus the discounted value of future rewards. Mathematically,

\begin{equation}\label{eq:simple_q}
Q_i(S^i_t, a^i_t) = (1-\alpha) \overbrace{Q_i(S^i_{t}, a^i_{t})}^{\text{Current estimate}} + \alpha \left(\overbrace{r^i_t + \gamma \underset{a^i}{\max} {Q_i(S^i_{t+1}, a^i)}}^{\text{Updated estimate}} \right),
\end{equation}

\noindent where $\underset{a^i}{\max} \;{Q(S_{t+1}^i, a^i)}$ is the value of the best possible action at the state the agent transits after the action taken at its previous state-action pair; $\alpha$ is the learning rate; $\gamma$ is the discount factor adjusting the importance of future rewards; and the difference between the updated estimate and the current estimate is the \gls{td} error. During exploitation, the agent selects the action with the highest Q-value, aiming to maximize future rewards.


\subsection{\gls{ma-drl}: Algorithmic and multi-agent interaction}

The need to build a Q-table with a Q-value for each possible state-action pair limits the scalability to larger state or action spaces. To address this, \gls{rl} is extended to \gls{drl}~\cite{sutton2018reinforcement}, where Q-functions are approximated using \glspl{dnn}, enabling agents to handle larger and more complex state spaces efficiently. In \gls{drl}, each agent $i$ stores every experience tuple, $\text{SARS}:$ $(S^i_t, a^i_t, r^i_t, S^i_{t+1})$, for learning purposes. These experiences are then used to adjust the weights of the \gls{dnn} via \gls{sgd}. By continuously minimizing the difference between the predicted Q-values and the {received rewards and observed next state}, \gls{sgd} fine-tunes the network's weights, thereby enhancing the policy's performance over time.

The innovative aspect of our proposed \gls{ma-drl} algorithm lies in observing the impact of an action $a^i_t$ from the perspective of a packet $p$ with destination $d$, $p(d)$. When $p(d)$ is forwarded by an agent $i$ to another agent $j$, it transits from the state $S^i_t$ observed in $i$ to $S^j_{t+1}$ observed at $j$. This experience tuple $\text{SARS:}$ $(S^i_t, a^i_t, r^i_t, S^j_{t+1})$ with states observed in the interacting agents is then stored in an experience buffer $D$ that is then used to train a \gls{dnn} characterized by weights $\theta$ that learns the optimal routing policy (Fig.~\ref{fig:Phases}.b). The modified {Q-Learning} equation is described as follows:


\begin{equation}\label{eq:q}
Q_{i}(S^i_t, a^i_t;\theta) \leftarrow (1-\alpha) Q_{i}(S^i_t, a^i_t;\theta) + \alpha \left(r^i_t + \gamma \underset{a^j}{\max} Q_{j}(S^j_{t+1}, a^j;\theta)\right),
\end{equation}

\noindent where $Q_i(S^i_t, a^i_t;\theta)$ is the Q-value for the state-action pair at $i$ with weights $\theta$, and $\underset{a^j}{\max}Q_i(S^j_{t+1}, a^j;\theta)$ is the maximum value for every possible $a^j$ after observing $S^j_t$ at $j$. 
{The feedback needed by agents is minimal: First, the agent $i$ needs its neighboring agents congestion information in order to observe $S^i_t$. This vector of 16 values that can be updated periodically. Then, agent $i$ needs the new maximum possible value encountered at $p(d)$'s receiving agent, $j$, and the time spend on its queue in order to compute $r^i_t$ (Fig \ref{fig:Phases}.b). Therefore, only 2 numbers are needed for evaluating the action made.}


\subsection{Loss function and \gls{ddqn}}
\label{subsec:DDQN}

The experience tuples $\text{SARS}$ are used to train a \gls{dnn} referred to as the Q-Network $Q(\theta)$, with its parameter weights $\theta$. In the context of \gls{ddqn}, an additional network called the Target Network ($Q(\theta^-)$) is utilized alongside $Q(\theta)$ to stabilize the learning process by addressing the overestimation bias present in the standard DQN algorithm \cite{van2016deep}.

The objective of DDQN is to minimize the following loss function, which is based on the \gls{td} error:

{
\begin{equation} \label{eq:loss}
    L(\theta) = \mathbb{E}\left[ \left( r^i_t + \gamma \overbrace{\max_{a^j} Q_i(S_{t+1}^{j}, a^{j}; \theta^-)}^{\text{Target Network}} - \overbrace{Q_i(S^i_t, a^i_t; \theta)}^{\text{Q-Network}} \right)^2 \right],
\end{equation}
}

\noindent where $r^i_t$ denotes the reward, $\gamma$ is the discount factor, $a^i_t$ is the action taken {at the observed state $S^i_t$ by the sending agent $i$}, $S_{t+1}^{j}$ represents the next state observed at the receiving agent $j$, $\theta$ are the parameters of the Q-Network, and $\theta^-$ are the parameters of the Target Network. The key feature of DDQN is the separation of the action selection and evaluation. The action $a^i_t$ that maximizes the Q-value in the current state $S_{t}^{i}$ is selected using the current $Q(\theta)$, but the evaluation of this action's Q-value is performed using $Q(\theta^-)$. 

\gls{sgd} is applied to adjust the parameters of the Q-Network, $\theta$, by minimizing the loss function defined in eq. (\ref{eq:loss}). The parameters of the Target Network $\theta^-$ are updated less frequently than those of the Q-Network to provide a stable target for the \gls{td} error calculation. The update of the Target Network is performed through hard updates, where the parameters of the Q-Network are copied to the Target Network at fixed {time} intervals.

\subsection{\gls{ma-drl} continual learning}

One key feature of the proposed \gls{ma-drl} routing is its adaptability to fluctuations in traffic patterns and unavailable links, by each agent \emph{observing} its {one-hop} neighbouring agents congestion and link state. For a complete solution, it is important to consider the long-term evolution of the system and non-stationarities due to, e.g., non-functional nodes, traffic variations, or architectural changes. The latter is important during the roll-out phase of these networks, as dense constellations are often launched in batches over time. For a \emph{continual {decentralized} learning} that adapts to such evolution, we divide the learning into two phases, offline and online, as presented next.


\begin{algorithm}[t]
 \caption{Offline {phase}}
 \begin{algorithmic}[1]
    \renewcommand{\algorithmicrequire}{\textbf{Initialize:}}
    \REQUIRE Initialize $Q(\theta)$, $Q(\theta^-)$ parameters randomly
    \REQUIRE Initialize memory {buffer} $D_g$
    
    \FOR { $t=1,2,\dotsc,T$}
        \STATE {Get next packet $p_j \in \mathcal{P}$ arriving at satellite $i$ with destination $d$}
        \STATE Update $\mathcal{E}_S$ and $\mathcal{E}_G$ using {the \emph{greedy}} matching
        \IF {$d \in \mathcal{E}_{G}$}
            \STATE Deliver $p_j(d)$ to destination {$d$}
            \STATE $r^i_t = w_1 \cdot r_\text{q} + w_2 \cdot r_\text{r} + w_3 \cdot r^\star$
        \ELSE
            \STATE Observe sate {$S^i_t$}
            \IF {$r\sim U(0,1) < \varepsilon(t)$}
                \STATE Select a random action $a^i_t$
            \ELSE
                \STATE Select $a^i_t = \max_{a^i} {Q(S^i_{t}}, a^i; \theta)$
            \ENDIF
            \STATE $r^i_t = w_1 \cdot r_\text{q} + w_2 \cdot r_\text{r} + w_3 \cdot r^\star$
            \STATE Store transition $(S^i_t, a^i_t, r^i_t, S_{t+1}^{j})$ in $D_g$
            \STATE Perform \gls{sgd} to minimize $ L(\theta) $ with $\text{eq.}~(\ref{eq:loss})$
            \STATE Update $\theta$ as in eq.(~\ref{eq:q})
            \IF {t \%T == 0}
                \STATE Update Target Network parameters: $\theta^- \leftarrow \theta$
            \ENDIF
        \ENDIF
    \ENDFOR
 \end{algorithmic}
 \label{alg:offline_learning} 
 \end{algorithm}


\section{Offline phase} \label{sec:Offline_learning}

The offline phase (Fig.~\ref{fig:Phases}.c) initializes and trains a global model with a diverse and extensive dataset. Once trained, its knowledge can be transferred individually to each satellite in the online phase (Fig.~\ref{fig:Phases}.d).
The global model is characterized by two \glspl{dnn}, as in Section~\ref{subsec:DDQN}: the global Q-Network, $Q_g(\theta)$, and the global Target Network, $Q_g(\theta^-)$, with parameter weights $\theta$ and $\theta^-$ randomly initialized. 
Both networks share an identical architecture. The input layer comprises 28 neurons, each one corresponding to a different field of the observed state. Following the design in \cite{mnih2013playing}, there are 2 fully-connected hidden layers, each with 32 neurons. The output layer consists of 4 neurons, with each neuron representing the value of one of the 4 possible actions. 
In order to collect the most diverse dataset, all the satellites around the globe store their experience tuples $\text{SARS}$ in a global experience buffer $D_g$. The training process involves using the experience stored in $D_g$, applying \gls{sgd} to minimize the loss function  in (\ref{eq:loss}), and updating gradually $Q_g(\theta)$. $Q_g(\theta^-)$ receives hard updates periodically every $T$ {iterations}. We use an $\varepsilon$-greedy policy to balance exploration (i.e., select a random action) and exploitation (i.e., choose the action that maximizes the expected reward) through the probability $\varepsilon$. Specifically, $\varepsilon$ is given by:
\begin{equation}
    \varepsilon(t) = \varepsilon_{\text{min}} + (\varepsilon_{\text{max}} - \varepsilon_{\text{min}}) \cdot e ^{\kappa \cdot \frac{t}{n_{g}^2}},
    \label{eq:epsilon}
\end{equation}
where $\varepsilon_{\text{max}}$ and $\varepsilon_{\text{min}}$ are the maximum and minimum values, respectively, $\kappa$ is a factor to control the decay rate, $n_{g}$ is the number of active gateways and $t$ is the time index. $\varepsilon$ decays with time following a negative exponential behaviour. This decay is smoothed as more gateways are active since the agent needs more time to learn more paths between gateways.

Upon the arrival of packet $p$ with destination $d$ at time $t$, the global model gets the state $S^i_t$ observed by agent $i$ and chooses an action $a^i_t$ to forward $p(d)$ to one of the neighbours. The reward $r^i_t$ is received, and the state encountered by $p(d)$, ($S^j_{t+1}$), is observed by the receiving agent $j$. The tuple experience {$\text{SARS}:$ $(S^i_t, a^i_t, r^i_t, S^j_{t+1})$} is stored at $D_g$. Notice that even when the global buffer stores experiences around the globe, the routing decisions are {decentralized}, using only the local information {at the observed state}. The decision-making mechanism oscillates between leveraging $Q_g(\theta)$ with the observed state to predict action $a^i_t$ and adopting random action selection. This exploration-exploitation trade-off is essential for optimizing the performance in the \gls{rl} context. This approach enables the agent to adapt to novel scenarios, discovering not only the optimal trajectory but also alternative routes. Initially, as the exploration rate $\epsilon$ is high and the \gls{dnn}s' parameters -- $\theta$ and $\theta^-$ -- are initialized randomly, the agent makes random routing actions. As time goes by, the agent learns and the exploration rate decreases exponentially, favoring the exploitation strategy over exploration. The implementation of the algorithmic of the offline exploitation phase is outlined in Algorithm \ref{alg:offline_learning}. 

Notice that the global experience buffer $D_g$ is enriched through the aggregation of experience data from each individual agent around the globe. The resulting model is global in the sense that it can be applied at any position and network/congestion condition, as it has learnt routing strategies for each situation due to the exploration routing policy from the perspective of every satellite. Then, it can be uploaded at each satellite-agent for the online phase. 



\section{Online phase} \label{sec:Online_learning}

Once the global model is trained to route packets between gateways under different conditions in a {decentralized} way, the models are uploaded on-board {of each satellite-agent} and the online phase starts. 
In this phase, each satellite $i$ is an autonomous agent with its own Q-Network, $Q_i(\theta)$ and  Target Network, ($Q_i(\theta^-)$), whose parameter weights $\theta$ and $\theta^-$  are taken from the global model {Q-Network} $Q_g(\theta)$ (Fig.~\ref{fig:Phases}.d). 
The experience tuples $\text{SARS}$ collected locally by each agent $i$ are stored in their respective buffer $D_i$ and used to keep training their local model. 
As agents now train with local data and learn region-specific traffic patterns, their knowledge begins to diverge. 
This divergence requires further knowledge sharing among the moving agents to converge again to a global model. We propose two strategies, one based on updates sharing between the satellite and its first neighbour ahead (a.k.a. model anticipation) and the other one on cluster-based \gls{fl}. In both cases, we aim at minimizing the overhead to the network by sharing local models to be aggregated.


\subsection{Model Anticipation}

The model anticipation phase (Fig.~\ref{fig:HighLevel}.a) leverages transfer learning between agents without the need for a global or cluster \gls{ps}, and minimizes the communication overhead by having model exchange only between pairs of satellites. In this option an agent $i$ in an orbital plane $o$ sends its Q-Network, $Q_i(\theta)$, to the adjacent agent, $i-1$, which is moving towards $i$'s former position. I.e., 

\begin{equation}
Q^{i-1}(\theta) \leftarrow \frac{1}{2} \left( Q_i(\theta) + Q_{i-1}(\theta) \right)
\end{equation}

This mechanism ensures that $i-1$ inherits the learned traffic flow patterns from $i$, eliminating the need for $i-1$ to independently learn these patterns. This method speeds up the learning process, as $i$ does not need to wait to find a solution independently and improves learning efficiency by reducing the number of adjustments $i-1$ must make to its local \glspl{dnn} for an effective solution. The application of this method as the satellites move in the orbital plane allows the adaptation to short-term changes in the environment. For example, for an orbital plane of period $T_o$ and $S_o$ satellites, the updates could be done every $T_o/S_o$ seconds. Then, the long-term divergences can be adjusted with the second option. In any case, the choice of the frequency of updates depends on the changes in the environment.


\subsection{Cluster \gls{fl}}

The second approach uses \gls{fl}~\cite{matthiesen2023federated} and is illustrated in Figs.~\ref{fig:HighLevel}.b. and ~\ref{fig:HighLevel}.c. We leverage the stability and higher bandwidth of intra-plane \glspl{isl} by defining the agents within the same orbital plane $o$ as a learning cluster $C_o$. 
The \gls{ps} acts as the global update scheduler~\cite{razmi2024scheduling} that decides when and how to execute a new round of communication for the updates. 

In each communication round, each agent $i$ within $C_o$ shares its own Q-Network, $Q_i(\theta)$. 
This is sent to its neighbor agent within $C_o$, $i+1$, 
which does the partial aggregation of the Q-Networks, denoted as $Q^*_i(\theta)$.
This aggregation is forwarded to the next agent $i+2$, and the recursive operation is repeated until all the agents in the cluster have participated and the $C_o$'s $Q^o(\theta)$ is given by the total aggregation 


{
\begin{equation}
Q^o(\theta) \leftarrow \sum_{i \in C_o} \frac{Q^*_i(\theta)}{|C_o|},
\end{equation}
}

\noindent where $|C_o|$ is the number of satellites in $C_o$. $Q^o(\theta)$ is then sent to the \gls{ps}. For simplicity, we assume the \gls{ps} to be located in a \gls{geo} satellite  with continuous connectivity with at least one satellite within each cluster\footnote{The impact of discotinuous connectivity in the aggregation process has been  studied in ~\cite{matthiesen2023federated} for a general application of \gls{fl}, and therefore applicable to the specific problem of routing}. The \gls{ps} computes the global model $Q_g(\theta)$ by aggregating all the cluster models and distributes it back to the clusters. This is done using \gls{fedavg}~\cite{mcmahan2017communication}.

The different agents' models within a cluster are then updated synchronously using a cluster update scheduler, while the different cluster models are updated synchronously using the global update scheduler.

This is an efficient way of keep training the satellites together, converging to a cluster solution without sharing the training information and minimizing the use of communication resources. Each satellite is updated with the aggregated cluster model that adjusts very fast to changes in the environment while saving energy as they agents learn from other's knowledge too, reducing the models divergence and required local training.


\subsection{Discussion on continual learning}

Coming back to Fig.~\ref{fig:Phases}, we conclude this section with insights of the practical use of the proposal, with the possibility of combining cluster and global \gls{fl} and model anticipation. Initially, the ground offline learning parameterized in the global Q-Network and Target Network ensures the establishment of a robust starting model, crucial for the subsequent online learning phase.
As the online phase starts each satellite $i$ in orbital plane $o$, now an autonomous agent equipped with its own \glspl{dnn} and $Q_i(\theta)$ and $Q_i(\theta^-)$, starts training with local data. As the satellites move, they receive frequent periodic updates to be aggregated to the current model from the predecessor (model anticipation).  
In addition, \gls{fl} is applied at another periodicity: Cluster $C_o$ updates within each orbital plane $o$ are aggregated much less frequently, and global updates across the entire constellation, needed for homogenization, occur even less often. This hierarchical approach ensures that, while each satellite benefits from localized learning adjustments, the overall network aligns to broader non-stationarities, maintaining consistency and optimizing network-wide performance. The framework makes each satellite both a contributor and beneficiary of a continually evolving knowledge base, while providing energy efficiency and minimizing the use of computation and communication resources, as each satellite has to process less data and share only the \glspl{dnn} weights.


\section{Performance evaluation} \label{sec:results}

\noindent\textbf{Framework and setup.}
Results were obtained by a simulator developed in Python. Our baseline scenario considers the \emph{Kepler} constellation design, with $M=7$ orbital planes at heights $h_m = 600$ km and $N_m = 20$ satellites per orbital plane, as illustrated in Fig.~\ref{fig:ISLs_Kepler}. Additionally, we consider three more constellations: (1) the \emph{Iridium Next} constellation (Fig.~\ref{fig:ISLs_Iridium}), with $M=6$, $h_m=780$ km and $N_m=11$; (2) the \emph{OneWeb} constellation (Fig.~\ref{fig:ISLs_OneWeb}), with $M=36$, $h_m=1200$ km and $N_m=18$; and (3) the \emph{Starlink} orbital shell at $h_m=550$ km (Fig.~\ref{fig:ISLs_Starlink}), with $M=72$ and $N_m=22$. The three first constellations follow a \emph{Walker star} architecture, while the Starlink shell follows a \emph{Walker delta} architecture~\cite{leyva2022ngso}.
Without loss of generality, there are up to $8$ transmitting and receiving gateways at ground positions around the globe, mostly taken from the existing KSAT network\footnote{https://www.ksat.no/services/ground-station-services/}. 
Specifically, we locate the gateways in Malaga (Spain), Los Angeles (US), Port Louis (Mauritius), Vardø (Norway), Nuuk (Greenland), Nemea (Greece), Azores (Portugal), and Bangalore (India). 
In the results, the simulator takes the first $|\mathbb{G}|$ from the sorted list above. 
The communication parameters used for the simulations are as follows. The transmission power is $10$\,W for the satellites and $20$\,W for the gateways. The carrier frequencies are $20$\,GHz for downlink, $30$\,GHz for uplink, and $26$\,GHz for \gls{isl}. We consider parabolic antennas of diameter $33$\,cm at the gateways and of $26$\,cm at the satellites. The system bandwidth for all the links is $W=500$\,MHz. The packet length is $B=64.8$\,kbits, as defined for DVB-S2~\cite{dvb_s2}. Unless otherwise indicated, the traffic load is set to  $\ell=1$. 
In the observed state, the granularity factor is set to $\sigma = 20$, resulting in neighbors and destination relative latitudes and longitudes falling within the ranges $[-4.5, 4.5]$ and $[-9, 9]$, respectively. The current satellite latitude and longitude are within the ranges $[0, 9]$ and $[0, 18]$, respectively. The encoded congestion values $C_{j,k}$ are defined to be within the range $[0, 10]$.


\noindent\textbf{Benchmarks.}
We compare the performance of our proposed \gls{ma-drl} algorithm to a source routing approach that use Dijkstra's algorithm~\cite{dijkstra1959} to find the \emph{Shortest path} to the destination. Specifically, the weights of the edges are set to be the inverse of the data rate, namely $w_{i,j}=1/R(i,j)$. This is a traditional routing approach that leads to choosing  routes with high data rate links. 
Notice that the \emph{Shortest path} algorithm has real time information about all the \gls{lsatc} (e.g., full knowledge), and represents the optimal, genie-aided solution if the network has no congestion and is limited only by the propagation times. In practice, it is impossible  to have real time information about the whole \gls{lsatc} (links and buffer status) due to the overhead caused by feedback messages and the inherent propagation times delays. On the other hand, \gls{ma-drl} has only 1 hop neighboring information at every hop, which is a realistic approach. 
We also compare the \gls{ma-drl} to the \emph{Q-routing} algorithm developed in our previous work~\cite{soret2023q}, where Q-Tables are used for routing instead of \glspl{dnn}. 
It is important to emphasize that the Q-learning method uses Q-Tables for the routing in front of \glspl{dnn} used in \gls{ma-drl}, without positional data and reduced queue details.


\begin{figure}[t]
    \centering
    \begin{subfigure}{0.48\textwidth}
        \centering
        \includegraphics[width=\linewidth]{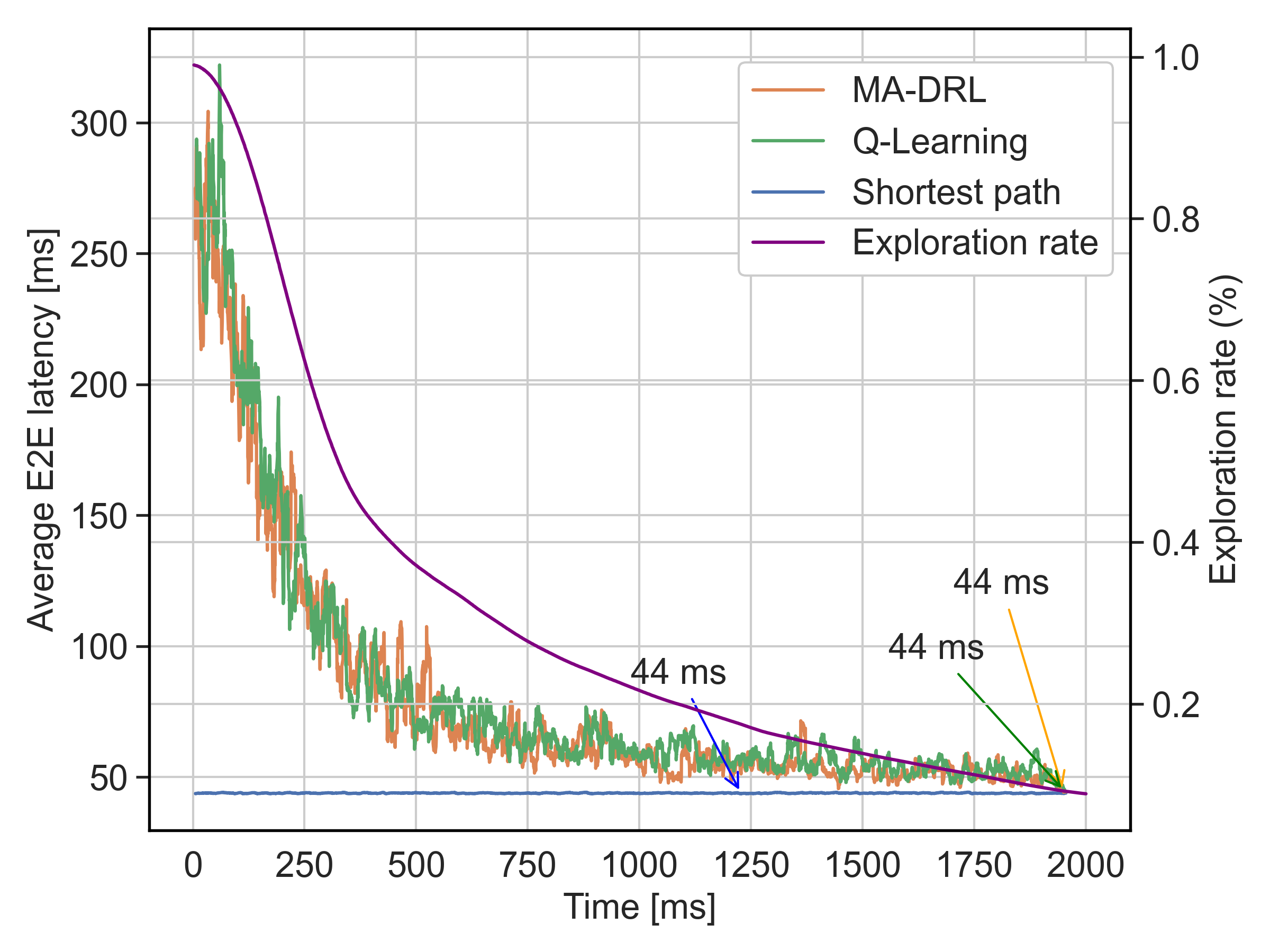}
        \caption{\emph{Kepler} constellation.}
        \label{fig:2GTsOfflineLatency_Kepler}
    \end{subfigure}%
    \hfill
    \begin{subfigure}{0.48\textwidth}
        \centering
        \includegraphics[width=\linewidth]{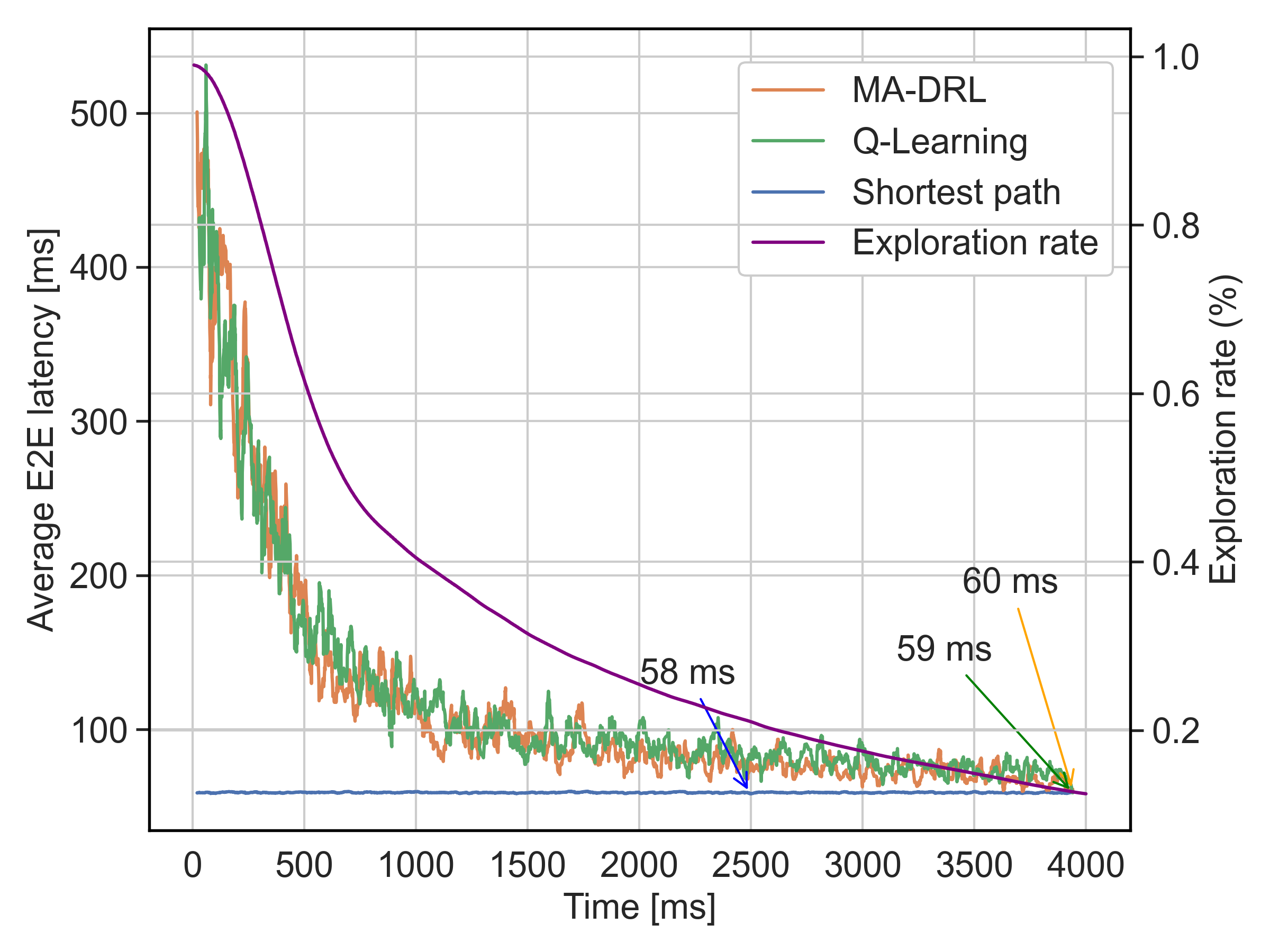}
        \caption{\emph{Iridium Next} constellation. 
        }
        \label{fig:2GTsOfflineLatency_Iridium}
    \end{subfigure}%
    \hfill
    \begin{subfigure}{0.48\textwidth}
        \centering
        \includegraphics[width=\linewidth]{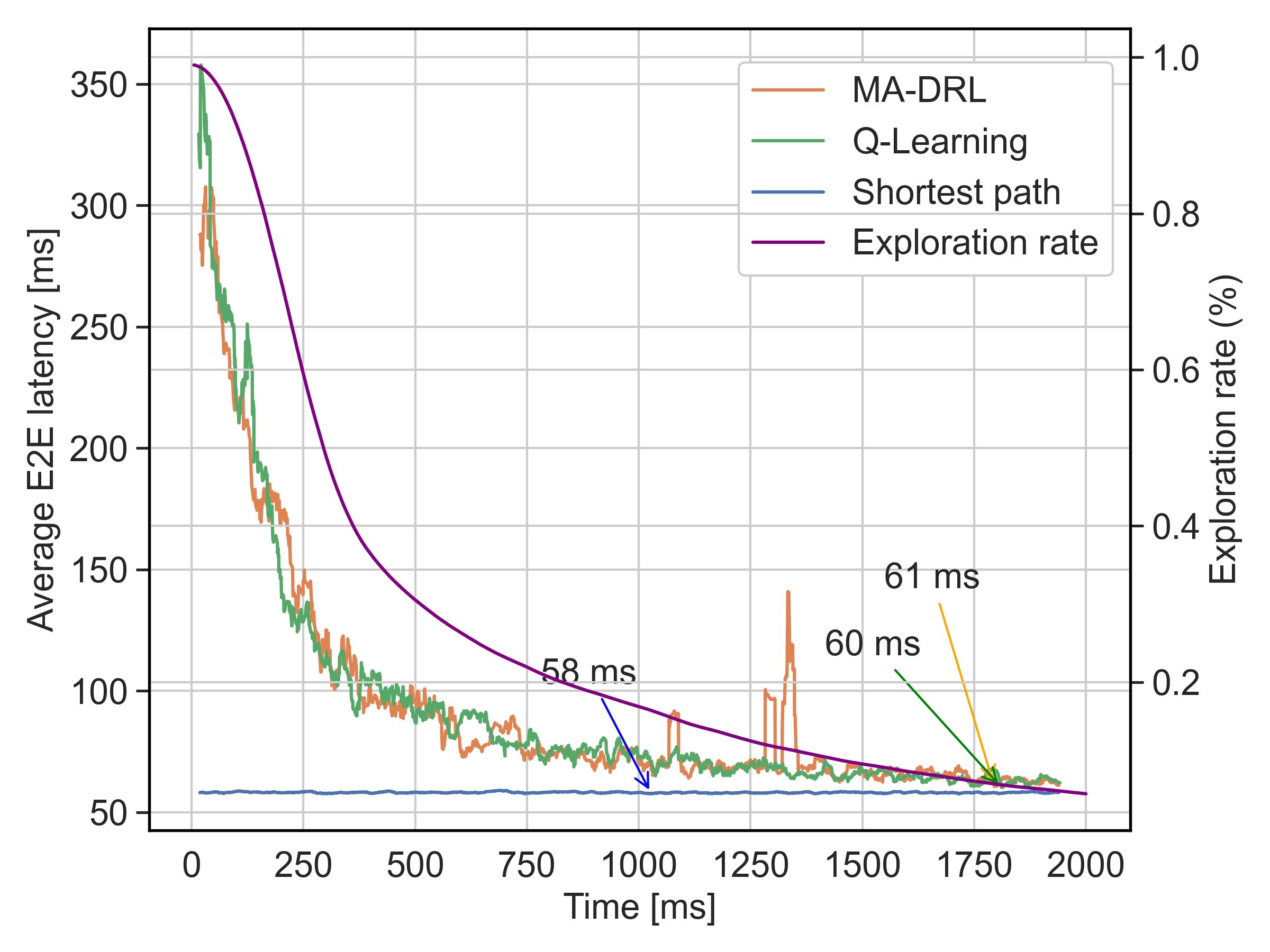}
        \caption{\emph{OneWeb} constellation. 
        }
        \label{fig:2GTsOfflineLatency_OneWeb}
    \end{subfigure}
    \hfill
    \begin{subfigure}{0.48\textwidth}
        \centering
        \includegraphics[width=\linewidth]{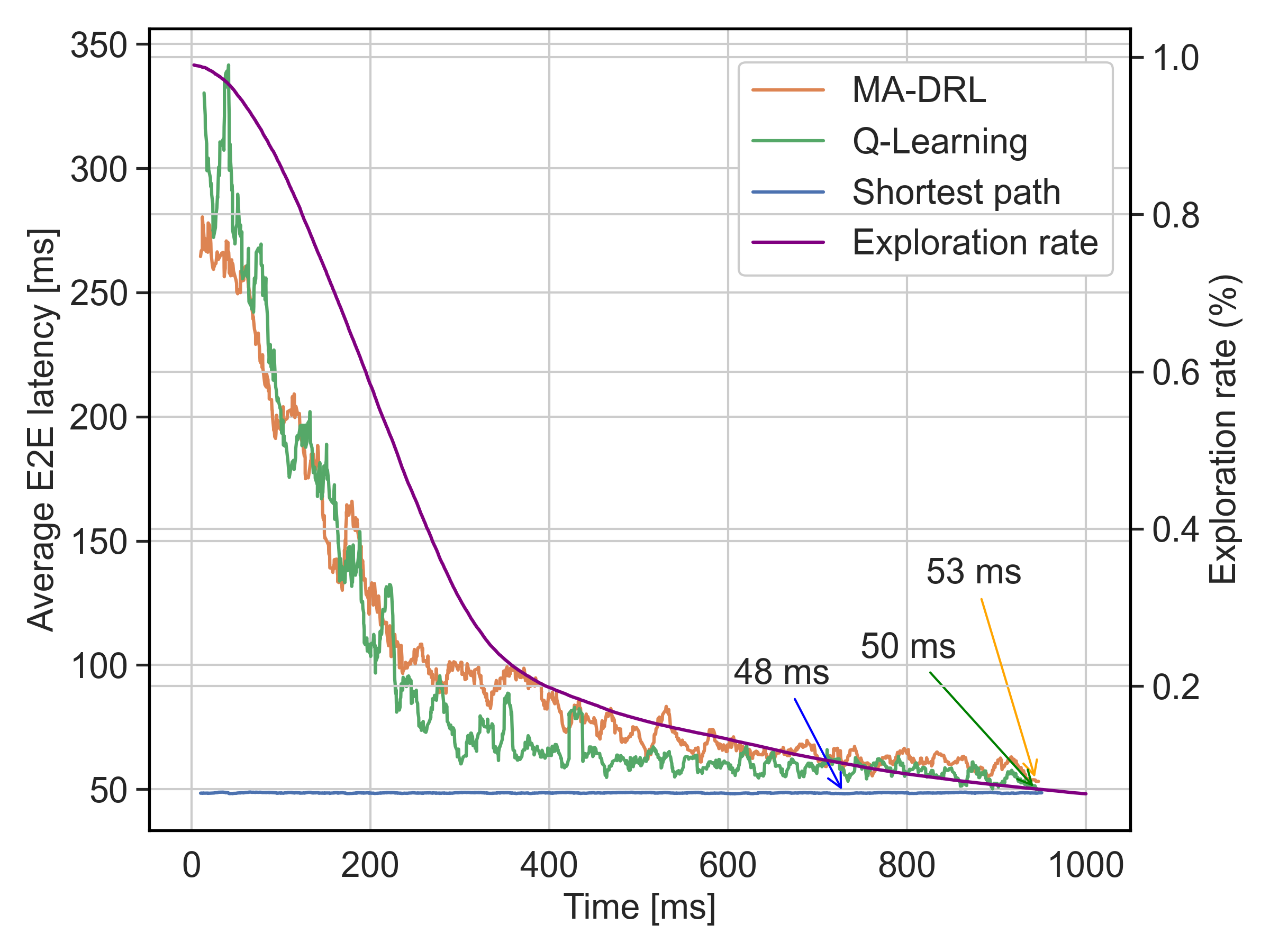}
        \caption{\emph{Starlink} constellation.
        }
        \label{fig:2GTsOfflineLatency_Starlink}
    \end{subfigure}
    \caption{\gls{e2e} latency and exploration rate versus time during the offline phase with 2 active gateways. Comparison of \gls{ma-drl} with the \emph{Q-routing} algorithm \cite{soret2023q} and the \emph{Shortest path} in the four defined constellation architectures at their initial deployment conditions.}
    
    \label{fig:constellationComparisons} \vspace{-0.6cm}
\end{figure}

\begin{figure}[t]
    \centering
    {\includegraphics[width=0.48\textwidth]{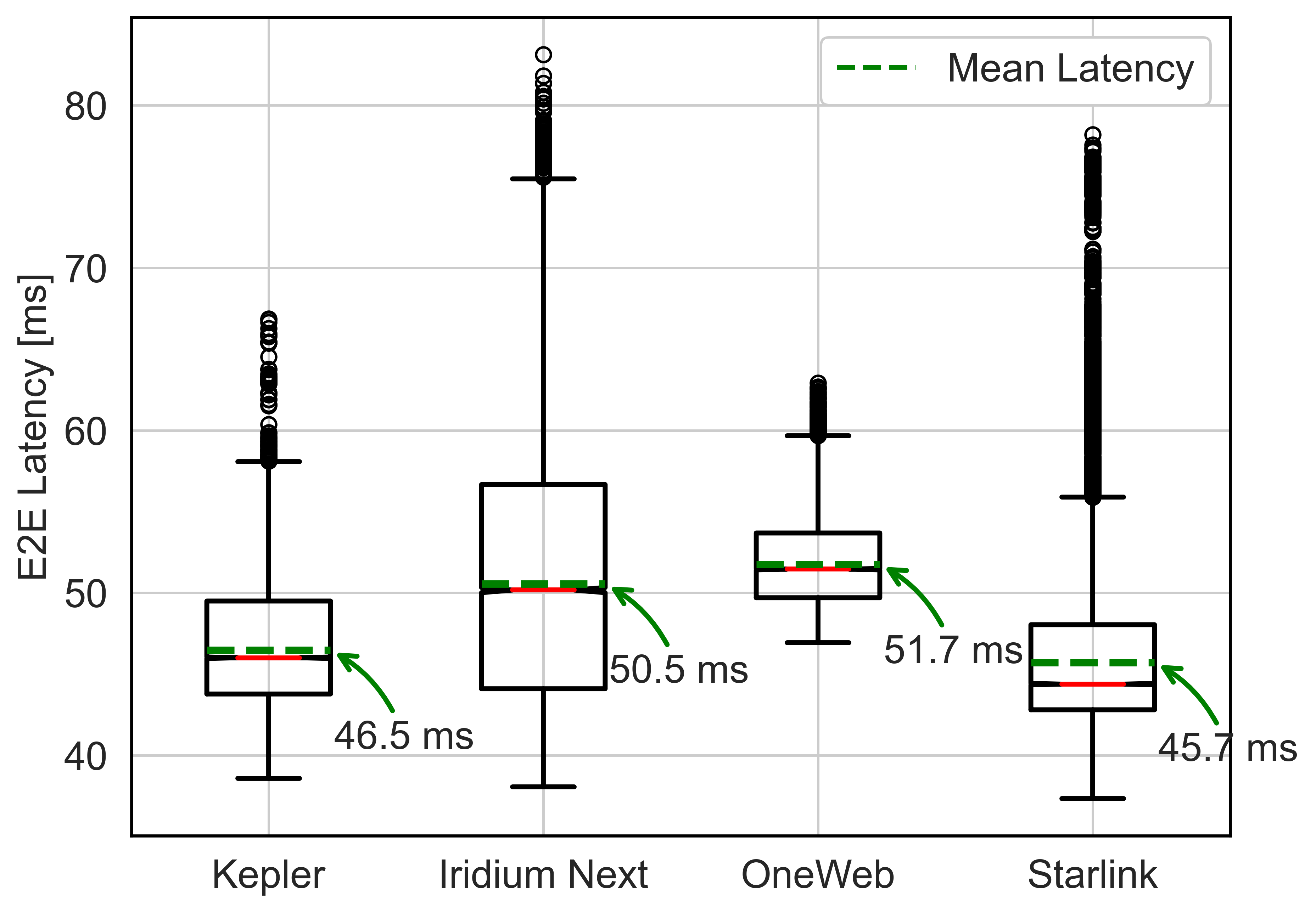}}
    \caption{Boxplot of the \gls{e2e} latency of the four constellation topologies with the \emph{Shortest Path} policy {after one orbital period is completed}.
    }
    \label{fig:boxPlot}\vspace{-0.5cm}
\end{figure}


\noindent\textbf{Offline phase.}
We consider 2 transmitting and receiving gateways at Malaga (Spain) and Los Angeles (US). We evaluate the behaviour of the four {described} constellations -- Kepler, Iridium, OneWeb, and Starlink -- with different satellite topology and densities (Fig~\ref{fig:ISLs}) . The corresponding \glspl{isl} that inter-connect the nodes are calculated with the \emph{greedy} matching algorithm. Fig. \ref{fig:constellationComparisons} shows the results, specifically the average E2E latency versus the packet arrival time in the left-axis, and the exploration rate versus packet arrival time in the right-axis. We compare the two learning solutions, \gls{ma-drl} and \emph{Q-routing}{~\cite{soret2023q}}, with the \emph{Shortest path}. The latter has no convergence or exploration, and in this evaluation without congestion its horizontal asymptote represents the optimal, genie-aided solution aiming at choosing the path with the highest data rate. During the exploration phase, 
the global model Q-Network, $Q_g(\theta)$, and target-Network, $Q_g(\theta^-)$, weights, $\theta$ and $\theta^-$, are initialized randomly, as well as the Q-Tables for the \emph{Q-routing}. Coupled with a high exploration rate $\varepsilon$, the global model initially tends to make random routing actions at each satellite. 
This behaviour disappears as $\varepsilon$ decreases: The global model learns first sub-optimal paths and then converges to the \emph{Shortest path}. This behavior emerges as a consequence of the global model's optimization process, aimed at maximizing its reward acquisition. 
The convergence time between the \gls{ma-drl} algorithm proposed in this paper and the \emph{Q-routing} algorithm are similar in all cases. For the \emph{Kepler} and \emph{OneWeb} constellations, convergence occurs within 2 seconds, whereas  for \emph{Iridium Next}  it occurs within 4 seconds and within 1 second for \emph{Starlink}. The reason for this behavior is that  \emph{Iridium Next} has the least number of satellites and, therefore, the length of the paths and the amount of routing decisions per packet are  lesser than for the other constellations.  Because of this, the global model gathers less experience for each routed packet. On the other hand, the fastest convergence is observed with the densest constellation, \emph{Starlink}, as it collects the most experience for each routed packet.

We elaborate on the comparison of the four topologies in Fig.~\ref{fig:boxPlot}, where the distribution of the E2E latency is depicted in a box plot {when the \emph{Shortest path} is applied}. In this figure, the constellation has moved to complete one orbital period in $96$ minutes, and satellite positions were updated at intervals of $15$ seconds. 

We observe that \emph{Kepler} and \emph{Starlink} obtain the smallest average latency, although the latter presents {more outliers}. 

\begin{figure}[t]
\centering
\begin{subfigure}{0.48\textwidth}
    \centering
    \includegraphics[width=\linewidth]{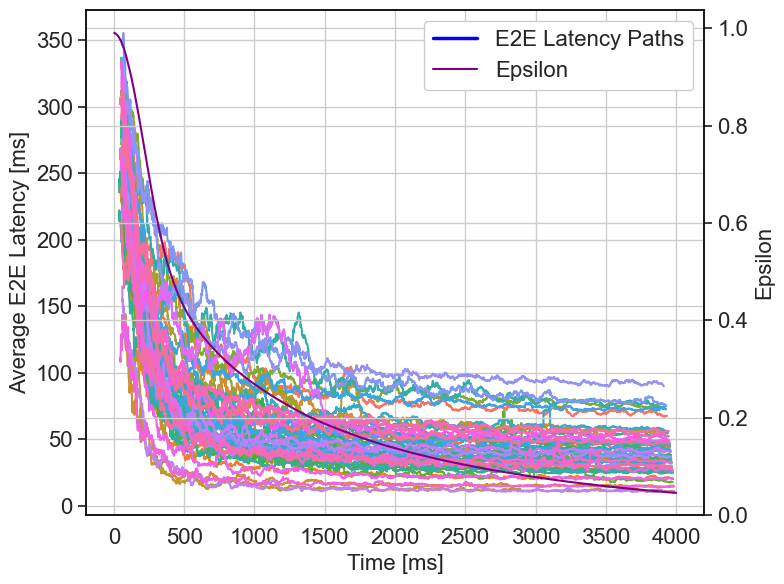}
    \caption{{Offline phase.}}
    \label{fig:8GTsOfflineLatency}
\end{subfigure}%
\hfill
\begin{subfigure}{0.48\textwidth}
    \centering
    \includegraphics[width=\linewidth]{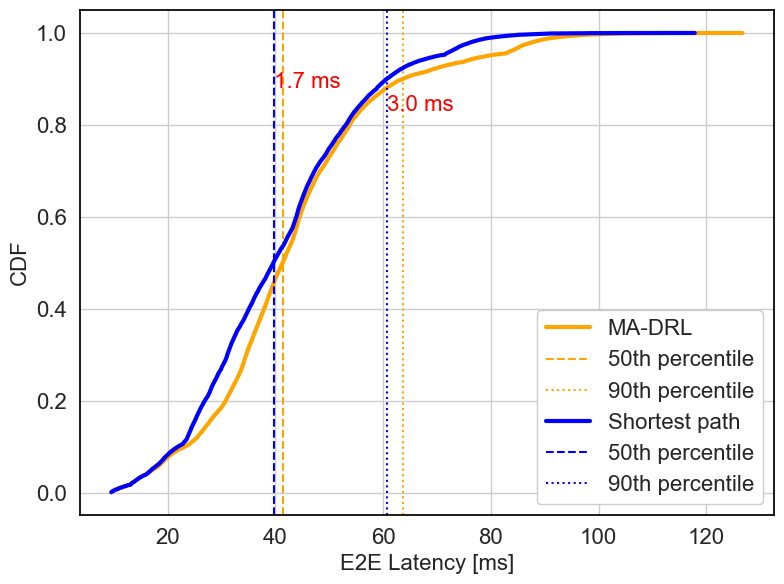}
    \caption{{Online phase.}}
    \label{fig:No_congestion_CDF}
\end{subfigure}
\caption{{\gls{e2e} Latency vs time} with 8 active gateways. (a) Offline phase: E2E latency versus packet arrival time (left-axis) and exploration rate (right-axis). (b) Online phase: Empirical \gls{cdf} of the latency. 
}
\label{fig:8GTs}
\end{figure}

After \gls{ma-drl} has demonstrated its ability to find optimal paths between 2 gateways, we test its learning capabilities {in a more complex environment} with higher load, i.e., 8 active gateways injecting traffic into the network, resulting in a total of $8 \times (8 - 1) = 56$ unidirectional paths and $\ell=0.5$. Notice that this is the only simulation where we set $\ell\neq1$, to make the network work at medium load. We do it for the \emph{Kepler} constellation. 
The latency results are shown in Fig.~\ref{fig:8GTsOfflineLatency}. The global model learns how to route packets {in a decentralized way from} 8 active gateways, each one sending packets to each other gateway through the \gls{lsatc}.
 \gls{ma-drl} has been trained for 4 seconds in the offline phase, as shown in Fig.~\ref{fig:8GTsOfflineLatency}, where it can be appreciated how the algorithm, from scratch and concurrently, finds paths between gateways, reducing the \gls{e2e} latency between them, {optimizing the received rewards.}

\noindent\textbf{Online phase.} 
The result of the offline phase (Fig~\ref{fig:8GTsOfflineLatency}) are two trained \glspl{dnn}: The global model Q-Network $Q_g(\theta)$, that weights $57$ KB and the global model Target Network $Q_g(\theta^-)$, that weights $27$ KB. These models are now uploaded on-board of each satellite. Fig.~\ref{fig:No_congestion_CDF} shows a comparison of the empirical \gls{cdf} of the \gls{e2e} latency of the packets when the routing is done following the \emph{Shortest path} policy, with full knowledge of the constellation, versus when the routing is done {decentralized} with local information, following the trained $Q_i(\theta)$ routing policy at each agent $i$.
It can be appreciated that the latency is very similar, i.e., the routes followed by the {decentralized} solution approximate the centralized approach with full knowledge of the constellation. 

Specifically, the 50th percentile gap is only 1.7 ms, whereas the 90th percentile gap is 3 ms, and the 95th percentile is 9.1 ms.

\begin{figure}[t]
\centering
\begin{subfigure}{0.48\textwidth}
    \centering
    \includegraphics[width=\linewidth]{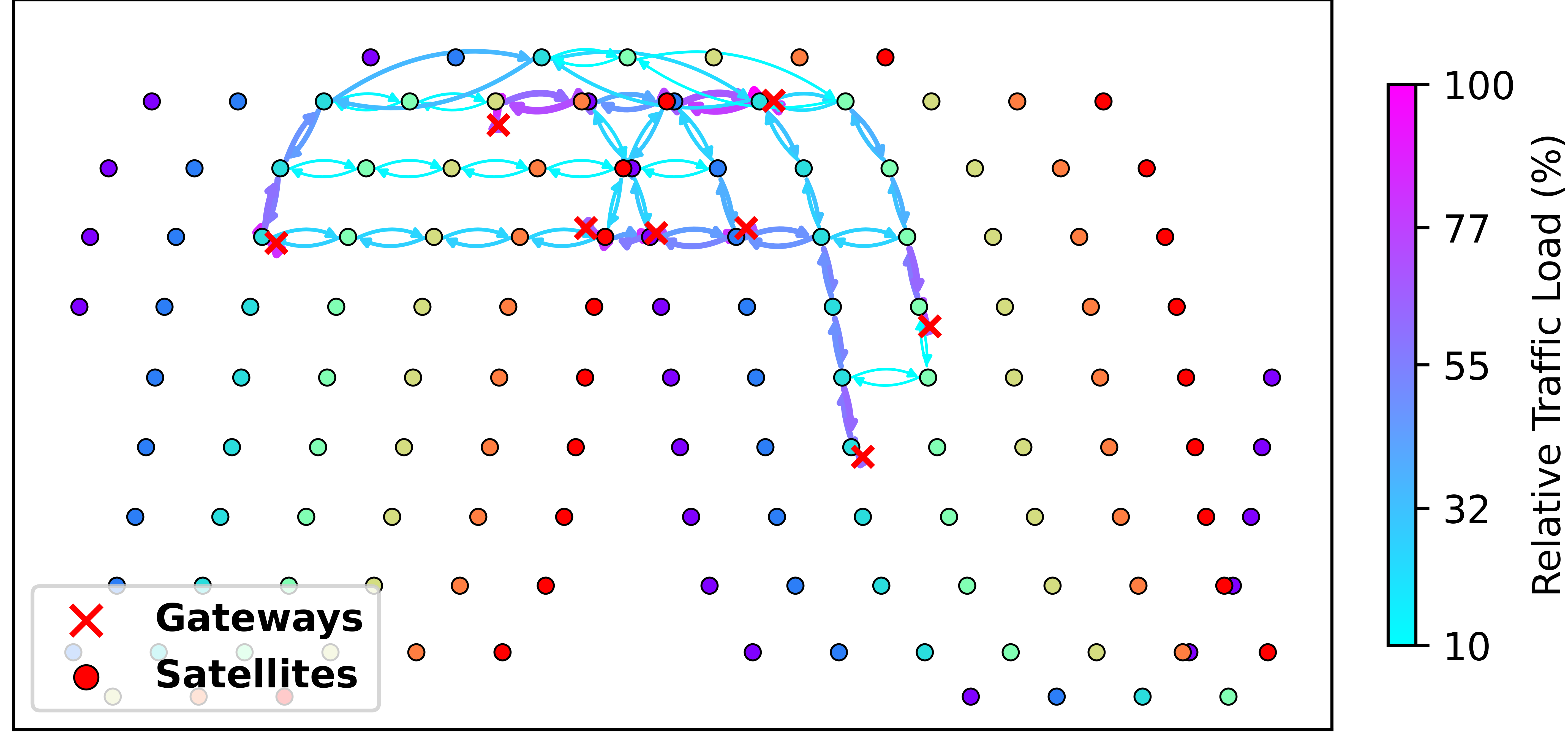}
    \caption{\emph{Shortest path} policy.}
    \label{fig:8GTsDataRateHeat}
\end{subfigure}
\hfill
\begin{subfigure}{0.48\textwidth}
    \centering
    \includegraphics[width=\linewidth]{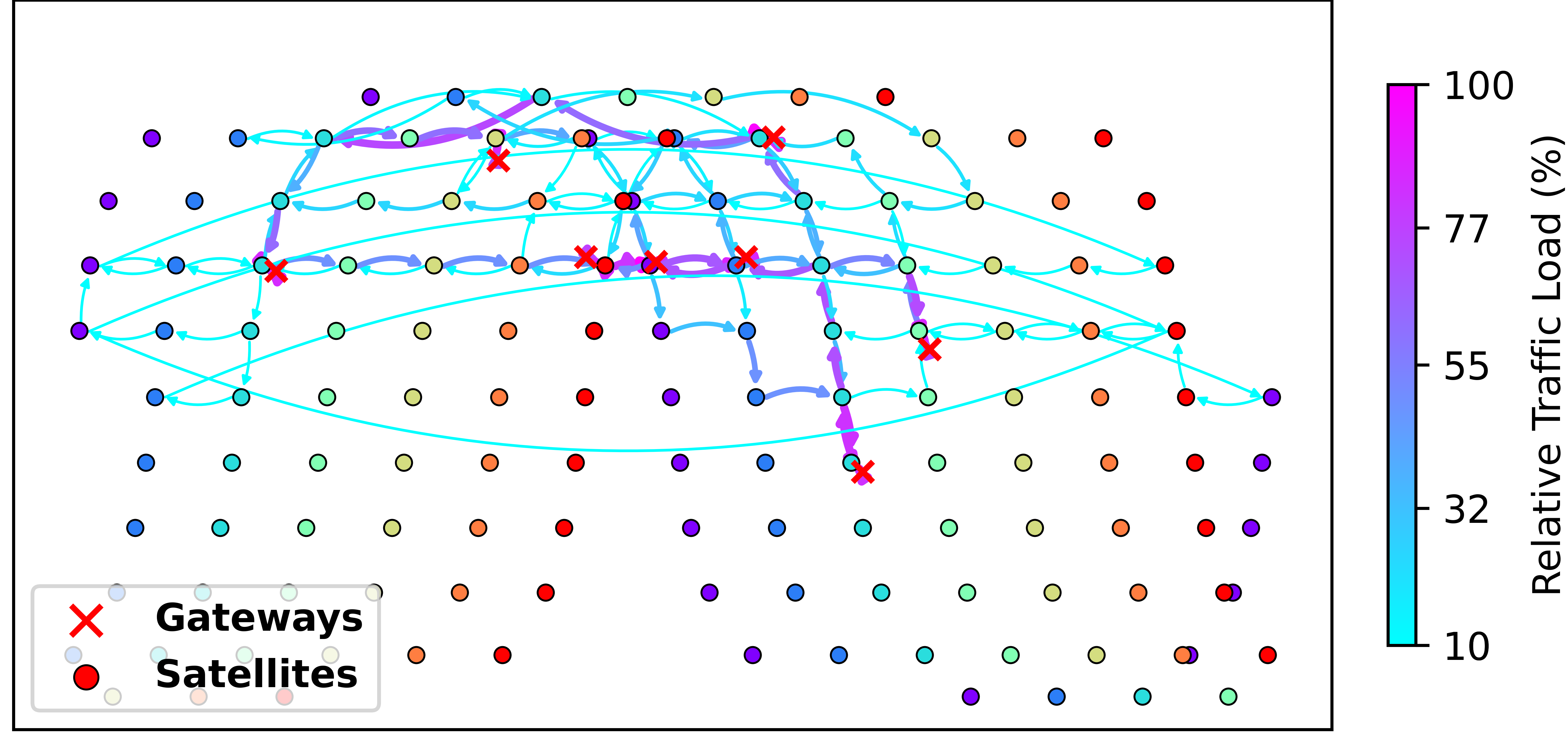}
    \caption{\gls{ma-drl} policy, online phase.}
    \label{fig:8GTsOnlineHeat}
\end{subfigure}
\caption{{Heat map of the preferred paths with $|\mathbb{G}| = 8$ and two policies: (a) \emph{Shortest path}; (b) \gls{ma-drl}. The 100$\%$ traffic load is given by the maximum number of packets that went through a single \gls{gsl}}}
\label{fig:8GTsHeat} \vspace{-0.6cm}
\end{figure}

Finally, Fig.~\ref{fig:8GTsHeat} shows a heat map of \gls{ma-drl} and \emph{Shortest path} with 8 gateways. The relative traffic load is represented to see the preferred paths. The \emph{Shortest path} algorithm calculates fixed paths which are concentrated in a part of the graph, while a considerable portion of the constellation supports none or little traffic. \gls{ma-drl} considers the queue congestion and exploits alternative paths with longer distances but less congested queues as the load increases. 


\begin{figure}[t]
\centering
\begin{subfigure}{0.48\textwidth}
    \centering
    \includegraphics[width=\linewidth]{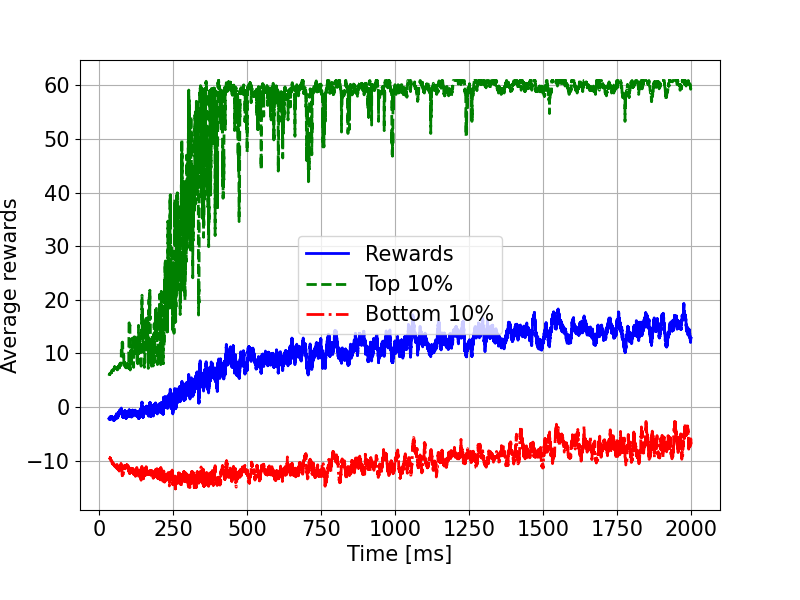}
    \caption{2 active gateways.}
    \label{fig:2GTsRewards}
\end{subfigure}
\hfill
\begin{subfigure}{0.48\textwidth}
    \centering
    \includegraphics[width=\linewidth]{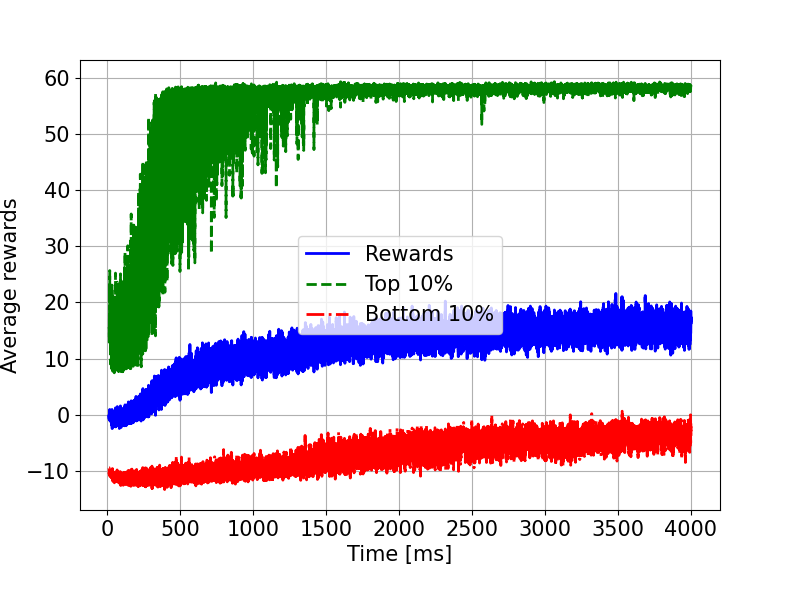}
    \caption{8 active gateways.}
    \label{fig:8GTsRewards}
\end{subfigure}%
\caption{Rewards over time of the offline model with (a) 2 active gateways; (b) 8 active gateways. The highest rewards are given after a delivery. 
}
\label{fig:Rewards}\vspace{-0.6cm}
\end{figure}

\noindent\textbf{Rewards.}
The averaged, top 10\% and bottom 10\% rewards for 2 and 8 gateways during the offline phase are shown in Fig.~\ref{fig:Rewards}. The values of $w_1, w_2$ and $w_4$ are {20, 20 and 5}, respectively. The deliver reward is 50, and the penalty for either making loops or choosing a not available link is -5. In the beginning, as the {model} is exploring, the rewards are lower. As the {model learns from experience, the rewards are higher and the \gls{e2e} latency decreases}. The top 10\% rewards correspond to the event of delivering the packet to the destination, as this action provides the highest reward.


\noindent\textbf{Movement.}

\begin{figure}[t]
    \centering
    {\includegraphics[width=0.48\textwidth]{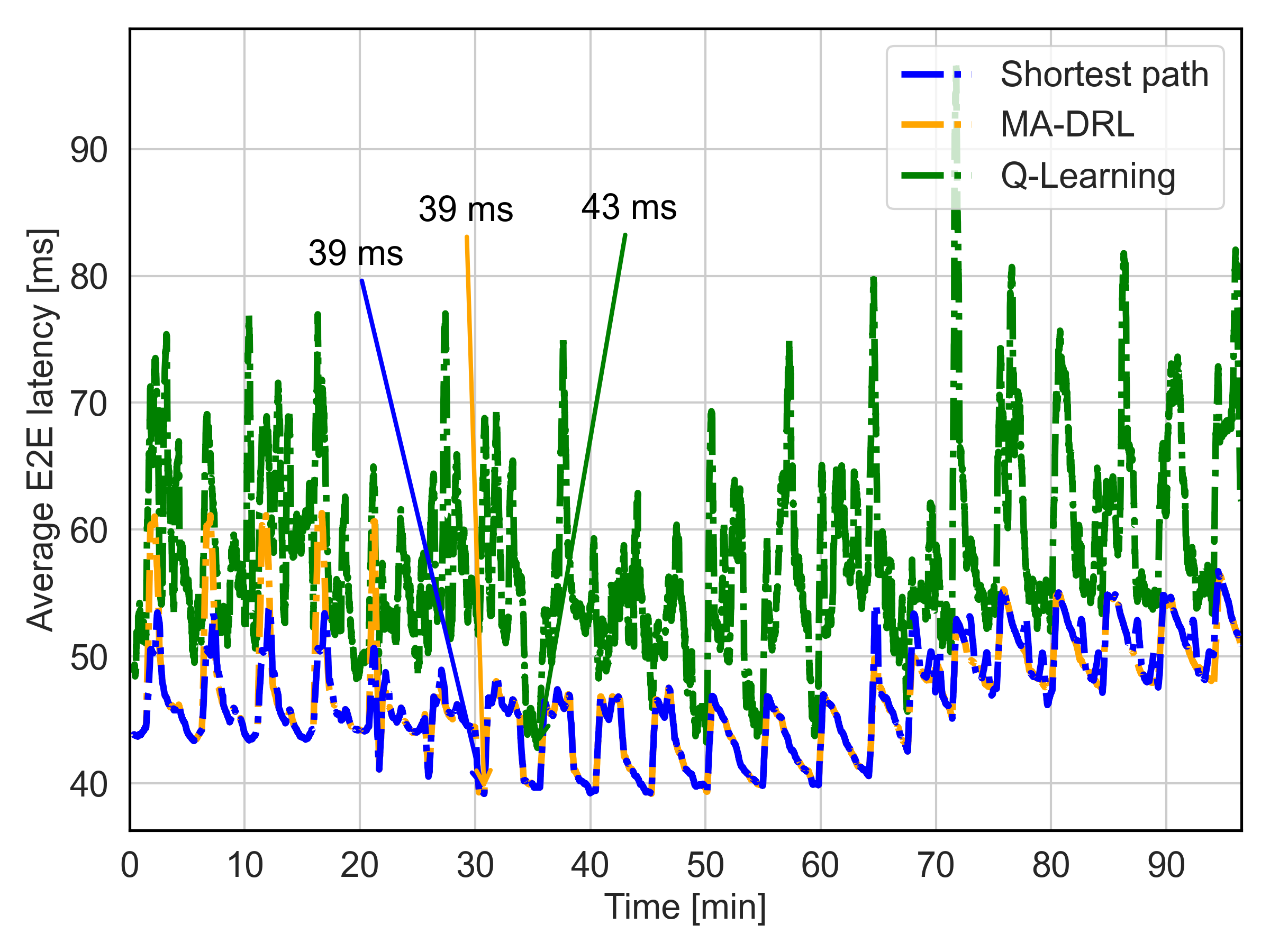}}
    \caption{Average \gls{e2e} latency over a complete orbital period of 96 minutes. Satellite positions were updated every 15 seconds. As satellites move, the topology changes and each gateway continuously links to its closest satellite, causing the path to change over time and fluctuations in the latency.
    }
    \label{fig:shifted2Gts}\vspace{-0.5cm}
\end{figure}

Next, we analyze the impact of the movement in the {online phase}. We consider 2 gateways and observe that the {agents} still find the optimal path, as shown in Fig.~\ref{fig:shifted2Gts}, without needing further exploration {and having the same latency as the \emph{Shortest path} (Which has real-time information of the full constellation) most of the time.} 
This adaptability comes from the fact that the global model, during the offline phase, has trained with local data with positional and congestion information in the state space, which makes its knowledge suitable for every agent at any position. {During the online phase, each agent has on-board a copy of the trained \glspl{dnn} from the global model. This allows them to adapt well to different observed states at different positions and congestion scenarios}. On the other hand, the \emph{Q-routing} algorithm~\cite{soret2020latency} needs to converge again and finds a sub-optimal route since it has no positional information and {must go} over the exploration phase {every time the topology changes.}

\noindent\textbf{Continual learning.}


\begin{figure}[t]
\centering
\begin{subfigure}{0.48\textwidth}
    \centering
    \includegraphics[width=\linewidth]{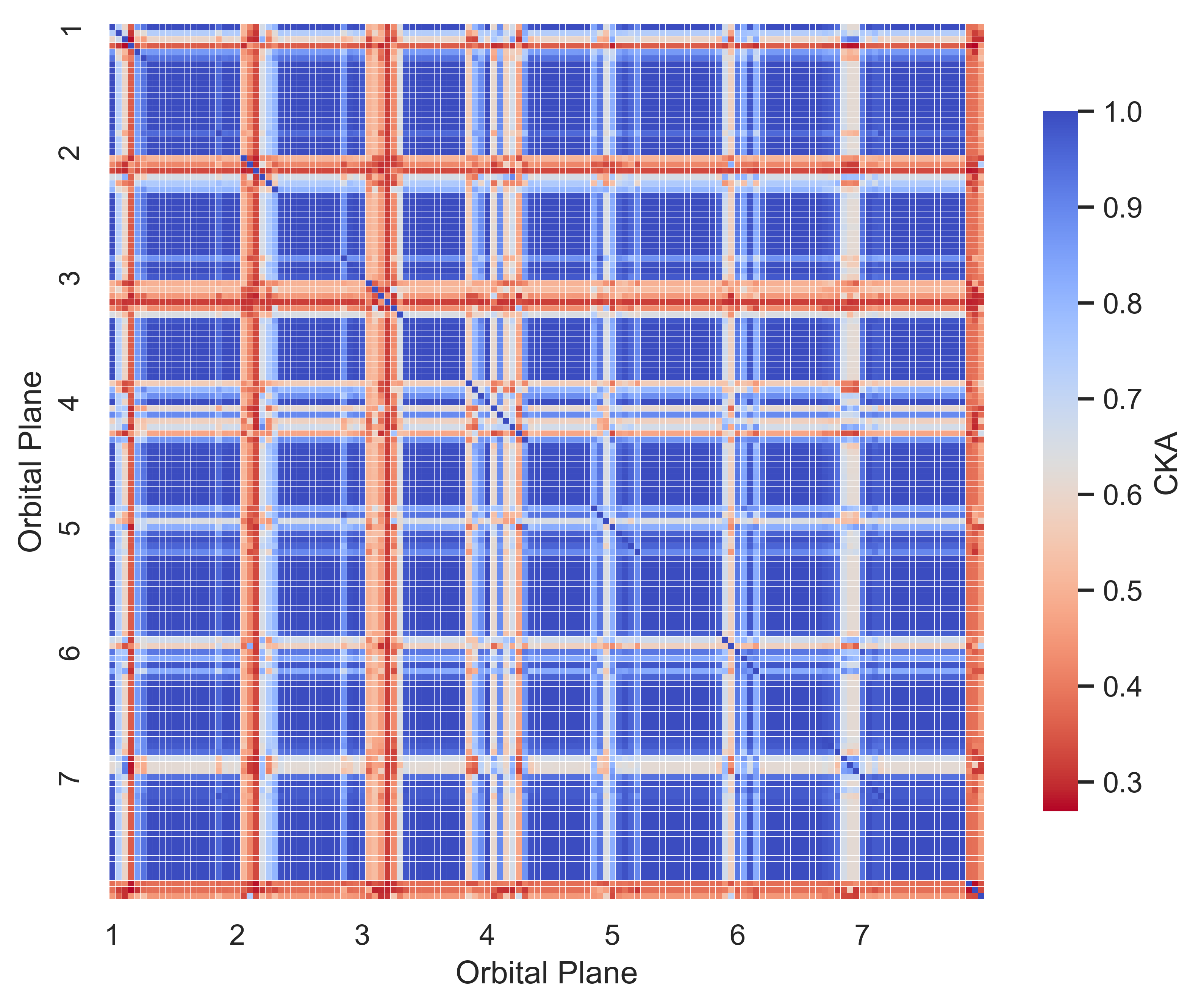}
    \caption{Without alignment}
    \label{fig:CKA_No_FL}
\end{subfigure}%
\hfill
\begin{subfigure}{0.48\textwidth}
    \centering
    \includegraphics[width=\linewidth]{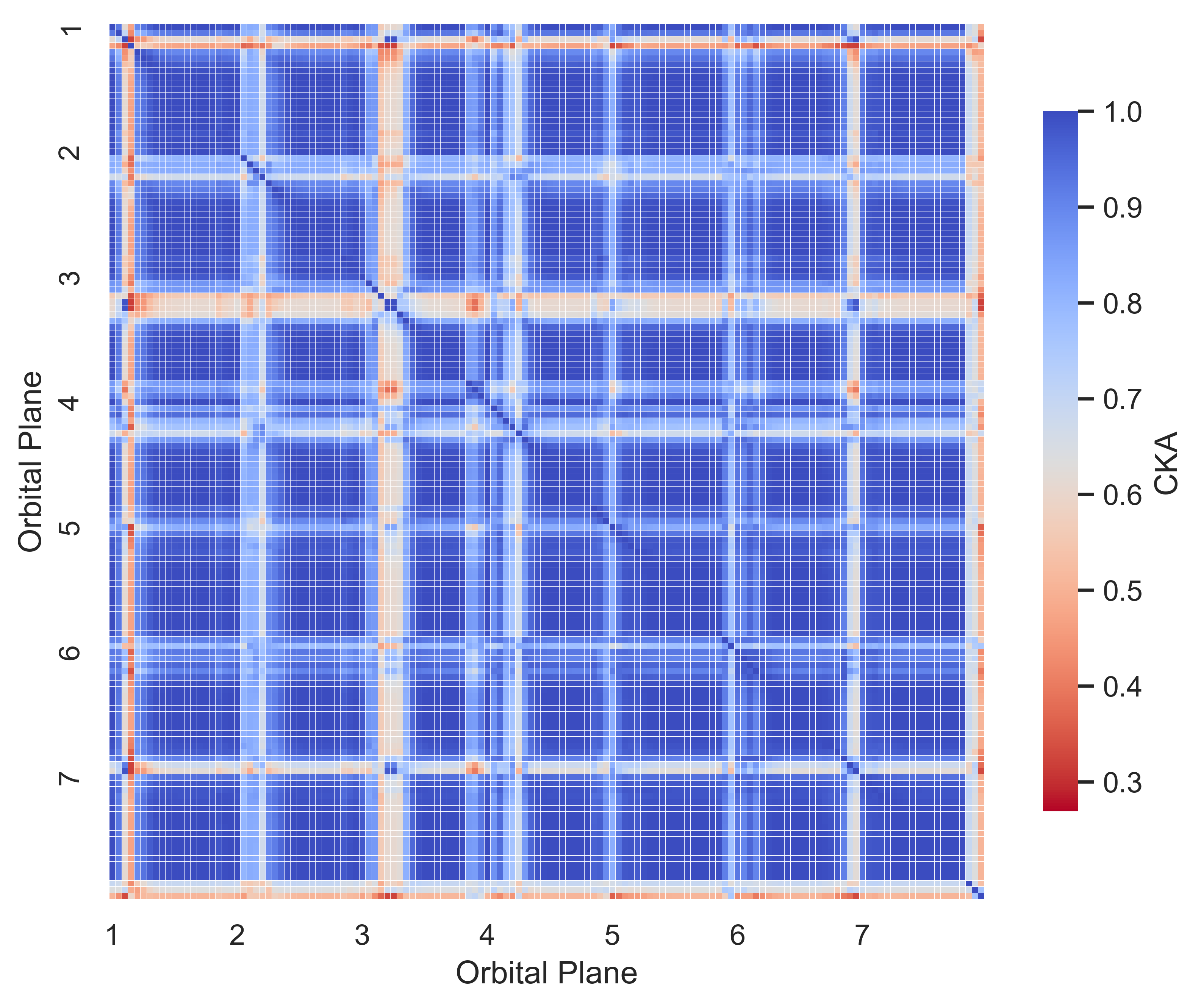}
    \caption{Model anticipation}
    \label{fig:CKA_Anticipation}
\end{subfigure}
\hfill
\begin{subfigure}{0.48\textwidth}
    \centering
    \includegraphics[width=\linewidth]{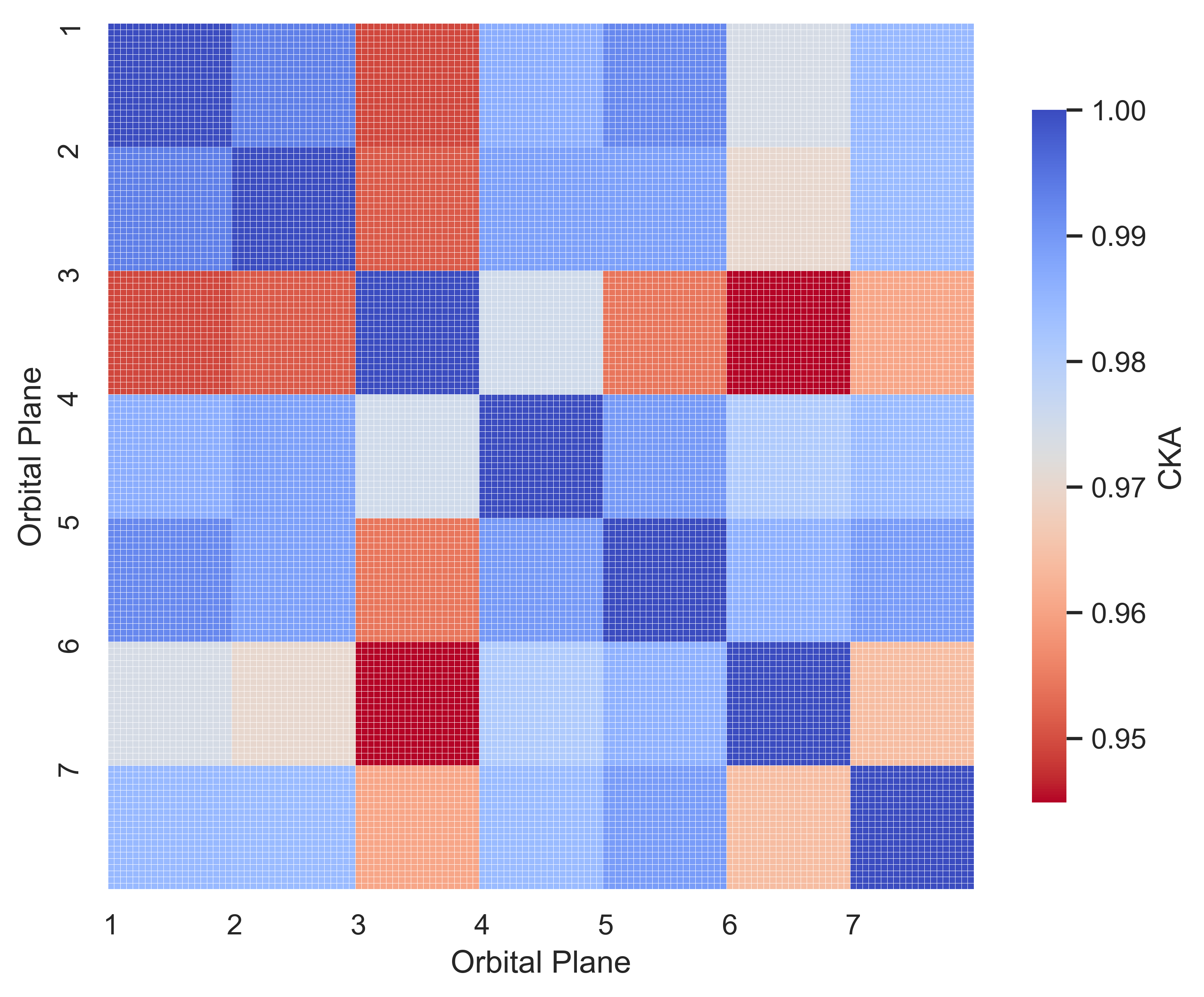}
    \caption{Orbital plane aggregation (FL)}
    \label{fig:CKA_FL}
\end{subfigure}
\hfill
\begin{subfigure}{0.48\textwidth}
    \centering
    \includegraphics[width=\linewidth]{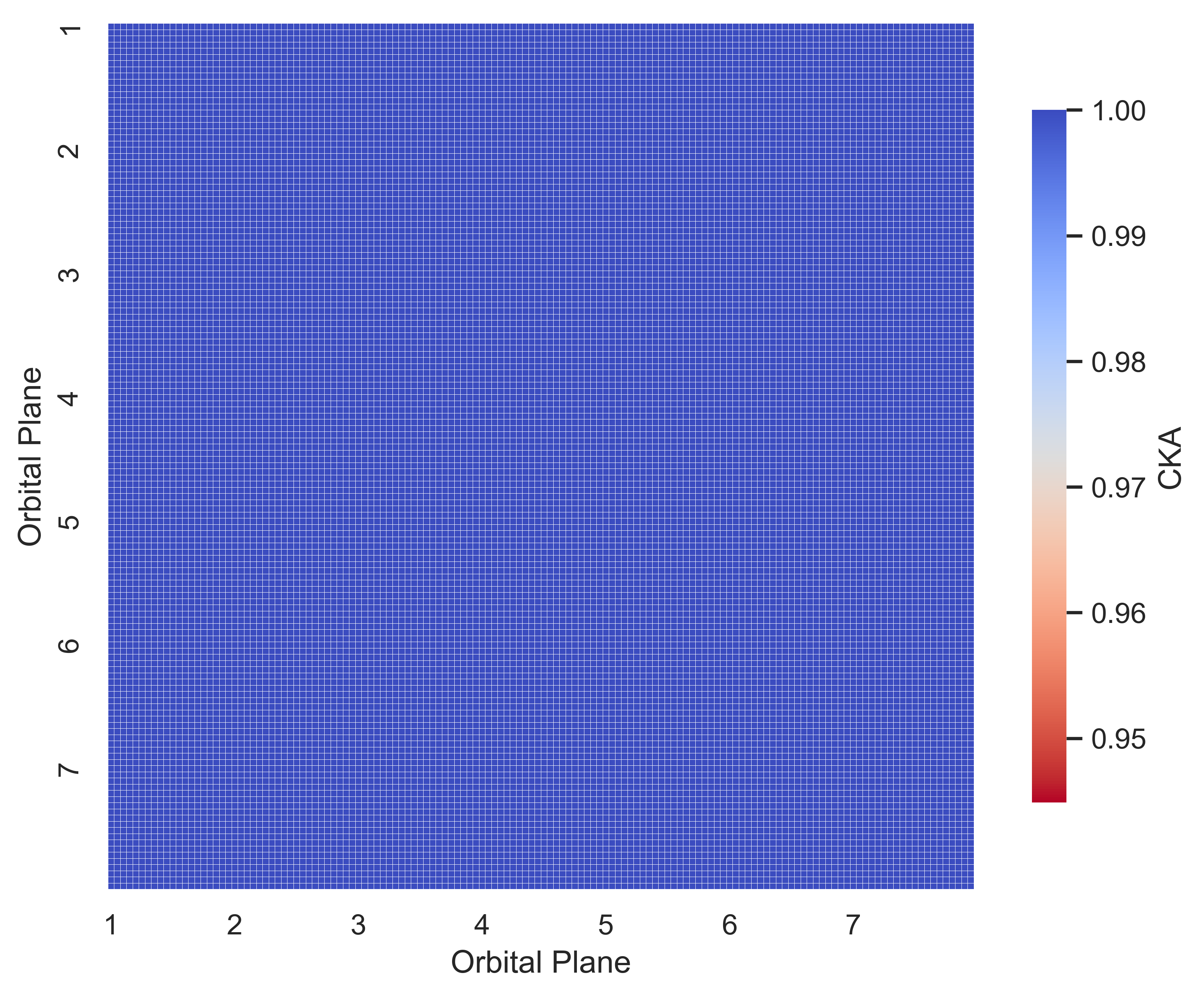}
    \caption{{Full aggregation (FL)}}
    \label{fig:CKA_FL_Full}
\end{subfigure}
\caption{{CKA computed values in different scenarios. Note that the color-map scale differs between the two upper figures and the two lower figures to enhance visualization clarity.}
} 
\label{fig:CKAs} \vspace{-0.6cm}
\end{figure}

\begin{figure}[t]
    \centering
    {\includegraphics[width=0.48\textwidth]{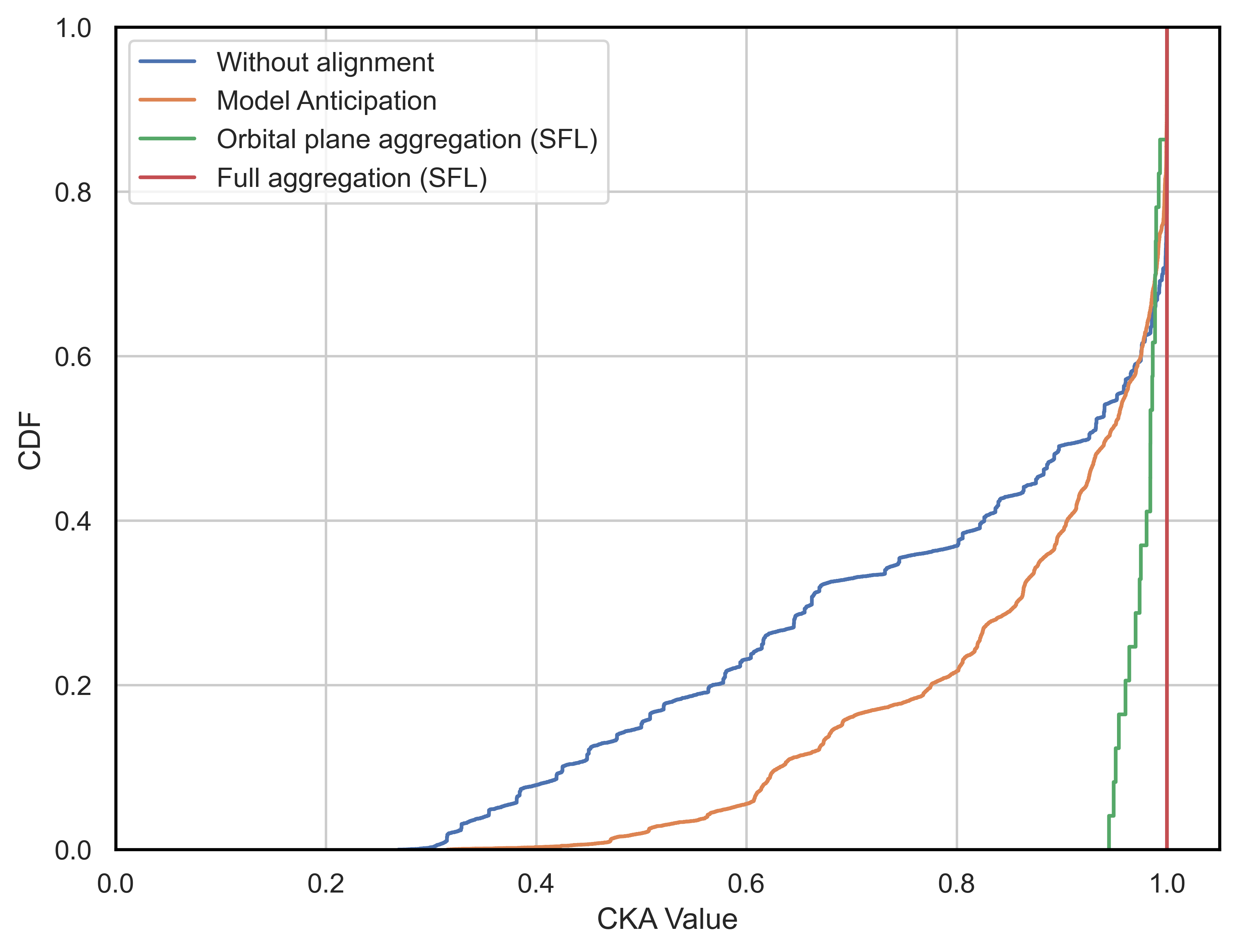}}
    \caption{{CKA CDF values for the scenarios presented in Fig.~\ref{fig:CKAs}. 
    }}
    \label{fig:CKA_CDF}
\end{figure}

Fig.~\ref{fig:CKAs} motivates the need for continual learning and demonstrates the performance of the proposed methods. We choose \gls{cka} as a similarity metric widely used to quantify the divergence among models {during the online phase} for diverse reasons. Given two sets of representations $X = x_1, ..., x_n$ and $Y = y_1, ..., y_n$ (models in our case), and assuming these representations are centered, the \gls{cka} between $X$ and $Y$ is given by \cite{kornblith2019similarity}
\begin{equation}
\text{CKA}(X, Y) = \frac{\text{tr}(X X^T)\cdot \text{tr}(Y Y^T)}{||X X^T||_F \cdot ||Y Y^T||_F}
\end{equation}

\noindent where $\text{tr(A)}$ denotes the trace of matrix $A$, $X X^T$ is the Gram matrix of $X$ and analogously for $Y$, and $||\cdot||_F$ denotes the Frobenius norm. A \gls{cka} value of $1$ indicates perfect alignment between the representations, while a value closer to $0$ indicates less alignment. Fig.~\ref{fig:CKA_No_FL} shows in a matrix form the value of \gls{cka} between each pair of satellites $i, j$ in the Kepler constellation with 8 gateways. {For this comparison, a realistic dataset was generated and input into each \gls{dnn} to evaluate their behavior.} After just one second, the average \gls{cka} value is 0.8. In this simulation, {as the traffic load is set to  $\ell=1$, the} network gets congested, and the increased queuing delay is the reason behind the divergence in the models: as satellites {observe a congested state and} experience longer delays in the predefined paths, they try with alternatives that incur in longer propagation delays but lower buffering. The diagonal corresponds to $i, i$ and therefore $\text{CKA}(x, x) = 1$. However, it is observed that many other pairs have diverged significantly, with values as low as $0.3$. The areas with less divergence correspond to the areas far from the gateways with paths that are never or rarely selected. With model anticipation, the alignment increases as shown in Fig.~\ref{fig:CKA_Anticipation}. The satellites send their Q-Networks to the successors and this results in a higher value of \gls{cka}, with an average improvement of {9.9\%}. However, a level of misalignment is still kept and therefore it is necessary to have a global update at the cluster and network level. This is the result in Fig.~\ref{fig:CKA_FL}, {with an average improvement of 22\%}.  Specifically, we show the first step of cluster aggregation, demonstrating that after one round of cluster communication the models are aligned at the cluster level. The second step is to aggregate all the cluster and reach $\text{CKA}(i, j) = 1\;\forall i, j${, as shown in Fig.~\ref{fig:CKA_FL_Full}}{, with the remaining average improvement of 24,6\% that results in the perfect alignment}. Fig.~\ref{fig:CKA_CDF} summarizes the results in an empirical \gls{cdf} of the \gls{cka}, showing that the initial model divergence (no alignment) is reduced with the model anticipation and fully achieved with \gls{fl}.


\section{Conclusions}
\label{sec:conclusions}

The proposed \gls{ma-drl} for packet routing in \glspl{lsatc} is a simple solution that has demonstrated promising results in terms of efficient learning, as it quickly converges to an optimal routing policy with minimal local status information at every hop, showcasing the effectiveness of the decentralized \gls{dnn}-based approach. Notably, with regard to adaptability, \gls{ma-drl} has not only learned the optimal path but also a set of alternative paths during the offline phase. This aspect is crucial in highly loaded \glspl{lsatc} where the queueing delay becomes critical. Thus, the agents, being aware of their neighbors' congestion status, can seamlessly switch to these alternative paths. Moreover, the online phase addresses a practical yet overlooked problem: the need for continual learning as the environment evolves and the local models uploaded at each agent start diverging. Our approach, with a short- and a long-term aggregation, reduces the disalignment at different scales, providing all the flexibility needed for a practical implementation that adapts to the specific needs of a space system.


\bibliographystyle{IEEEtran}
\bibliography{main}

\end{document}